\documentclass[11pt,eqsecnum,nofootinbib]{revtex4}
\usepackage{amsmath,amsfonts,bm}
\usepackage{hyperref}
\usepackage{url}

\newcommand{\be}{\begin{equation}}
\newcommand{\ee}{\end{equation}}
\newcommand{\bi}{\begin{itemize}}
\newcommand{\ei}{\end{itemize}}
\renewcommand{\le}{\left}
\newcommand{\ri}{\right}

\newcommand{\fanin}{\texttt{fan\_in}}
\newcommand{\Lb}{\lambda_b}
\newcommand{\LW}{\lambda_W}
\newcommand{\NTK}{\widehat{H}}
\newcommand{\dNTK}{\widehat{\mathrm{d}H}}
\newcommand{\ddNTKI}{\widehat{\mathrm{dd}_{\mathrm{I}}H}}
\newcommand{\ddNTKII}{\widehat{\mathrm{dd}_{\mathrm{II}}H}}
\newcommand{\E}[1]{\mathbb{E}\left[#1\right]}
\newcommand{\etanorm}{\eta_0}
\newcommand{\tra}{\widetilde{\alpha}}
\newcommand{\out}{f}
\newcommand{\A}{\mathcal{A}_t}

\begin{document}

\title{Meta-Principled Family of Hyperparameter Scaling Strategies}

\author{Sho Yaida\\
Meta AI\\
Meta Platforms, Inc.\\
Menlo Park, California 94025, USA\\
\texttt{shoyaida@meta.com}}

\begin{abstract}
In this note, we first derive a one-parameter family of hyperparameter scaling strategies that interpolates between the neural-tangent scaling and mean-field/maximal-update scaling. We then calculate the scalings of dynamical observables -- network outputs, neural tangent kernels, and differentials of neural tangent kernels -- for wide and deep neural networks. These calculations in turn reveal a proper way to scale depth with width such that resultant large-scale models maintain their representation-learning ability. Finally, we observe that various infinite-width limits examined in the literature correspond to the distinct corners of the interconnected web spanned by effective theories for finite-width neural networks, with their training dynamics ranging from being weakly-coupled to being strongly-coupled.
\end{abstract}
\maketitle

\section{Introduction}\label{sec:introduction}
As we  push the limits of AI-powered technologies -- e.g., to draw like (Dal\'{i}+WALL-E)/2~\cite{ramesh2021zero,ramesh2022hierarchical}, to speak like an omnilingual translator~\cite{costa2022no}, and to beat the imitation game like a human being~\cite{turing1950computing,srivastava2022beyond}
-- state-of-the-art models are inflating in size at a staggering pace. Anticipating this scaling trend to remain in vogue for the foreseeable future~\cite{hestness2017deep,kaplan2020scaling,hoffmann2022training}, it is imperative that we understand how to scale up models intelligently. Specifically, as deep-learning models grow both in width and depth, we'd like a principled way to adjust hyperparameters accordingly so that we won't have to restart the hyperparameter search from scratch each time models redouble.

Among a variety of hyperparameters used in practice, the most ubiquitous ones are those that control initialization distributions and learning rates. For these omnipresent hyperparameters, the following two scaling strategies have been suggested by theorists for wide and deep neural networks: one is the neural-tangent (NT) scaling strategy~\cite{jacot2018neural}, which naturally arose from the study of infinitely-wide neural networks as kernel machines~\cite{Neal1996,LBNSPS2017,MRHTG2018}; the other is the maximal-update (MU) scaling strategy~\cite{yang2021tensor}, which generalizes the mean-field limit of one-hidden-layer neural networks~\cite{mei2018mean,rotskoff2018trainability,sirignano2020mean1,chizat2018global} to deeper ones. In the wild today, unfortunately, neither strategies are followed systematically by practitioners: there are yet to be fair, methodical, and community-wide empirical tests as to which hyperparameter scaling strategies perform better for which dataset-task pairs with which network architectures and optimizers.\footnote{See, e.g., Ref.~\cite{yang2022tensor} for a test of the MU scaling strategy.}

In order to complement these putative empirical efforts, here we place various scaling strategies into a coherent theoretical framework.

In \S\ref{sec:pqr} we first review minimum ingredients for our study while introducing the general class of hyperparameter scaling strategies that we call $p_{\ell}q_{\ell}r$ scaling strategies, which are equivalent to the abc-parametrizations introduced in Ref.~\cite{yang2021tensor}.\footnote{For the purpose of our presentation, we find it tidier to adopt a convention slightly different from Ref.~\cite{yang2021tensor}. For those familiar with Ref.~\cite{yang2021tensor}, our conventions are related by
\begin{align}
2a_{1}+2b_{1}=&p_{1}\, ;\ \ \ 2a_{\ell}+2b_{\ell}=1+p_{\ell}\, \ \ \ \text{for}\ \ \ \ell=2,\ldots,L\, ,\notag\\
2a_{1}=&q_{1}\, ; \ \ \ 2a_{\ell}=1+q_{\ell}\, \ \ \ \text{for}\ \ \ \ell=2,\ldots,L\, ,\notag\\
c=&-r\, .\notag
\end{align}
We also identify the $L$-th layer as an output layer rather than attaching the extra $(L+1)$-th layer. More broadly, we follow the notations and conventions in Ref.~\cite{PDLT} and extend the results therein to study general hyperparameter scaling strategies.} In \S\ref{sec:principles}, we then derive a one-parameter family of hyperparameter scaling strategies from the principle of criticality for preactivations and the principle of learning-rate equivalence for the neural tangent kernel, that is, from the desiderata to avoid exploding/vanishing signals and to ensure that every layer contributes equally to learning~\cite{PDLT}. The resulting family, \eqref{eq:meta-scaling1}--\eqref{eq:meta-scaling5}, is parametrized by a metaparameter $s\in[0,1]$ and interpolates between the NT scaling at $s=0$ and MU scaling at $s=1$. In \S\ref{sec:seeing}, we further evaluate the degree of representation learning
by analyzing the first-order and second-order differentials of the neural tangent kernel
and find that their leading-order representation-learning effect is proportional to
\be\label{eq:main}
\gamma\equiv\frac{L}{n^{1-s}}\, ,
\ee
where $n$ is the width and $L$ is the depth of a neural network. Note that,  had we fixed the depth in scaling up the model, the MU hyperparameter scaling strategy with $s=1$ would be the only strategy that retains representation-learning capability for a large-scale neural network. However, it is critical to stress that, for any value of $s$, we can scale up the depth along with the width as $L\sim n^{1-s}$ to keep the emergent scale $\gamma$ fixed. As we detail in \S\ref{sec:engulf}, this leads to a richer taxonomy of representation-learning models in the infinite-model-size limit, and an even richer set of effective theories at finite size; some are governed by weakly-coupled training dynamics that can be analyzed by perturbation theory and the others are governed by strongly-coupled dynamics that call for nonperturbative approaches. Appendix~\ref{app:hierarchy} contains some further calculating, which presents the hierarchical structure of neural tangent kernel differentials for one-hidden-layer neural networks.

\section{Review and $p_{\ell}q_{\ell}r$ Scaling Strategies}\label{sec:pqr}
Deep neural networks are composed of many structurally similar operations stacked on top of each other. A prototypical architecture -- on which we focus all our concrete calculational effort throughout this note -- is a multilayer perceptron (a.k.a.~a fully-connected feedforward network), recursively defined by
\begin{align}
z_{i;\delta}^{(1)}=&b_{i}^{(1)}+\sum_{j=1}^{n_0} W^{(1)}_{ij}x_{j;\delta}\,  \ \ \ \text{for}\ \ \ i=1,\ldots,n_1\, ,\label{eq:MLP1st}\\
z_{i;\delta}^{(\ell+1)}=&b_{i}^{(\ell+1)}+\sum_{j=1}^{n_{\ell}} W^{(\ell+1)}_{ij}\sigma\le(z_{j;\delta}^{(\ell)}\ri)\, \ \ \ \text{for}\ \ \ i=1,\ldots,n_{\ell+1}\, ;\ \ell=1,\ldots,L-1\, ,\label{eq:MLPrec}
\end{align}
where $x_{j;\delta}$ are inputs, indexed by a vectorial index $j$ and a sample index $\delta\in\mathcal{D}$ in a dataset $\mathcal{D}$;  biases $b^{(\ell)}_i$ and weights $W^{(\ell)}_{ij}$ are learnable model parameters;  the single-variable function $\sigma(z)$ is called an activation function; and $z^{(\ell)}_{i;\delta}$'s are called $\ell$-th-layer preactivations, with $\out_{i;\delta}\equiv z^{(L)}_{i;\delta}$ being the network output for a sample $\delta$. In addition, here, the input dimension $n_0$ and the output dimension $n_L$ are fixed by the dataset and by the task while the hidden-layer widths $n_1,n_2,\ldots,n_{L-1}$ and the depth $L$ are \textit{architecture} hyperparameters one freely chooses. For the purpose of our calculations, we assume that the numbers of the hidden-layer neurons are all of similar order,
\be
n_1,n_2,\ldots,n_{L-1}\sim n\, ,
\ee
i.e., that they are all being increased simultaneously and uniformly in scaling up the model.

Before training starts, learnable model parameters need to be initialized in some way. Typically, the biases and weights are initialized identically and independently with mean-zero distributions -- such as normal distributions and uniform distributions -- whose covariances will be denoted as
\be\label{eq:init}
\E{b_{i_1}^{(\ell)}b_{i_2}^{(\ell)}}=\frac{1}{n^{p_{\ell}}}\le(C_b^{(\ell)}\delta_{i_1i_2}\ri)\, , \ \ \ \E{W_{i_1j_1}^{(\ell)}W_{i_2j_2}^{(\ell)}}=\frac{1}{n^{p_{\ell}}}\le(\frac{C_W^{(\ell)}}{n_{\ell-1}}\delta_{i_1i_2}\delta_{j_1j_2}\ri)\, ,
\ee
where $C_b^{(\ell)}$ and $C_W^{(\ell)}$ are order-one numbers -- \textit{initialization} hyperparameters -- assigned to each layer, and expectation $\E{\cdot}$ is over the initialization distributions.\footnote{An obligatory apology is in order for using the symbol $\delta$ both for the Kronecker delta and for the sample index; it should be clear from the context which are which.} In particular, the NT scaling strategy corresponds to setting $p_{\ell}=0$ which, as we shall see, results in order-one preactivations for all the layers. More generally, nonzero $p_{\ell}$'s would let us consider other possible scaling strategies with different powers of the typical hidden-layer width $n$.\footnote{Even more generally, one can pick different $p_{\ell}$'s for biases and for weights; we will here make them equal such that biases and weights in a given layer contribute at the same order.}

After initialization, model parameters $\theta_{\mu}\equiv\le\{b_{i}^{(\ell)},W_{ij}^{(\ell)}\ri\}_{\ell=1,\ldots,L}$ are typically trained by some gradient-based optimization algorithm. A prototypical optimizer -- again to which we pay all our concrete calculational attention -- is gradient descent, given by
\be\label{eq:update}
\theta_{\mu}(t+1)=\theta_{\mu}(t)-\eta\sum_{\nu}\lambda_{\mu\nu}\frac{d\mathcal{L}_{\A}}{d\theta_{\nu}}\Big\vert_{\theta=\theta(t)}\, ,
\ee
iterated from $t=0$ at which point the network is initialized. Here, $\mathcal{L}_{\A}$ is a loss function on the batch of samples $\A\subset\mathcal{D}$ at the $t$-th iteration; $\eta$ is a global learning rate; and $\lambda_{\mu\nu}$ is a learning-rate tensor that lets us control in more detail how the gradient on the $\nu$-th model parameter affects the update of the $\mu$-th model parameter. Typically, the learning-rate tensor is chosen to be diagonal,
\be
\lambda_{b_{i_1}^{(\ell)}b_{i_2}^{(\ell)}}=\frac{1}{n^{q_{\ell}}}\le(\lambda_b^{(\ell)}\delta_{i_1i_2}\ri)\, , \ \ \ \lambda_{W_{i_1j_1}^{(\ell)}W_{i_2j_2}^{(\ell)}}=\frac{1}{n^{q_{\ell}}}\le(\frac{\lambda_W^{(\ell)}}{n_{\ell-1}}\delta_{i_1i_2}\delta_{j_1j_2}\ri)\, ,\label{eq:train}
\ee
where $\lambda_b^{(\ell)}$ and $\lambda_W^{(\ell)}$ are order-one numbers -- \textit{training} hyperparameters -- assigned to each layer: in words, $\eta\lambda_b^{(\ell)}/n^{q_{\ell}}$ is the learning rate for the $\ell$-th-layer biases and $\eta\lambda_W^{(\ell)}/(n^{q_{\ell}}n_{\ell-1})$ is the learning rate for the $\ell$-th-layer weights.\footnote{Even more typically, the learning-rate tensor is set to be a Kronecker delta, $\lambda_{\mu\nu}=\delta_{\mu\nu}$, giving the same learning rate for all the model parameters. That is the so-called standard parametrization, which is  used most standardly in practice as of today but maybe not if you come from the theoretical future.} Here again, the NT scaling corresponds to setting $q_{\ell}=0$ and nonzero $q_{\ell}$'s allow more general scaling strategies. In the same
spirit, we can think of further scaling the global learning rate with $n$,
\be\label{eq:eta-scale}
\eta=n^r\etanorm\, ,
\ee
with an order-one number $\etanorm$.

In passing, we note that there is trivial gauge redundancy among the exponents $p_{\ell}$, $q_{\ell}$, and $r$. That is, for any real number $g$, we readily see that the transformation
\be\label{eq:gauge}
p_{\ell}\rightarrow p_{\ell}\, , \ \ \ q_{\ell}\rightarrow q_{\ell}+g\, , \ \ \ r\rightarrow r+g\, 
\ee
leaves the initialization~\eqref{eq:init} and the update equation~\eqref{eq:update} invariant. I.e., out of the $2L+1$ exponents introduced in this section, one linear combination has no physical consequence.

Finally, to introduce various dynamical observables of our interest, let us Taylor-expand the update for the network output, $\out_{i;\delta}=z_{i;\delta}^{(L)}$, in the update for the model parameters,
\be\label{eq:Schrodinger}
\theta_{\mu}(t+1)-\theta_{\mu}(t)=-\eta\sum_{\nu}\lambda_{\mu\nu}\frac{d \mathcal{L}_{\A}}{d \theta_{\nu}}(t)
=-\eta\sum_{\nu}\lambda_{\mu\nu}\sum_{j=1}^{n_L}\sum_{\tra\in\A}\frac{d \out_{j;\tra}}{d\theta_{\nu}}(t)\frac{d \mathcal{L}_{\A}}{d \out_{j;\tra}}(t)\, .
\ee
After doing just that and rearranging summations and dummy indices, we get
\begin{align}
\out_{i;\delta}(t+1)=&\out_{i;\delta}(t)-\sum_{j=1}^{n_L}\sum_{\tra\in\A}\eta H_{ij;\delta\tra}(t)\frac{d \mathcal{L}_{\A}}{d \out_{j;\tra}}(t)\, \label{eq:output-update}\\
&+\frac{1}{2}\sum_{j_1,j_2=1}^{n_L}\sum_{\tra_1,\tra_2\in\A} \eta^2\mathrm{d}H_{ij_1j_2;\delta\tra_1\tra_2}(t)\frac{d \mathcal{L}_{\A}}{d \out_{j_1;\tra_1}}(t)\frac{d \mathcal{L}_{\A}}{d \out_{j_2;\tra_2}}(t)\, \notag\\
&-\frac{1}{6}\sum_{j_1,j_2,j_3=1}^{n_L}\sum_{\tra_1,\tra_2,\tra_3\in\A} \eta^3\mathrm{dd}_{\mathrm{I}}H_{ij_1j_2j_3;\delta\tra_1\tra_2\tra_3}(t)\frac{d \mathcal{L}_{\A}}{d \out_{j_1;\tra_1}}(t)\frac{d \mathcal{L}_{\A}}{d \out_{j_2;\tra_2}}(t)\frac{d \mathcal{L}_{\A}}{d \out_{j_3;\tra_3}}(t)+\ldots\, ,\notag
\end{align}
where we introduced
\begin{align}
H_{i_0i_1;\delta_0\delta_1}\equiv&\sum_{\mu,\nu} \lambda_{\mu\nu}\frac{d\out_{i_0;\delta_0}}{d\theta_{\mu}}\frac{d\out_{i_1;\delta_1}}{d\theta_{\nu}}\, ,\label{eq:NTKdef}\\
\mathrm{d}H_{i_0i_1i_2;\delta_0\delta_1\delta_2}\equiv& \sum_{\substack{\mu_1,\nu_1\\ \mu_2,\nu_2}}\lambda_{\mu_1\nu_1}\lambda_{\mu_2\nu_2}\frac{d^2\out_{i_0;\delta_0}}{d\theta_{\mu_1}d\theta_{\mu_2}}\frac{d\out_{i_1;\delta_1}}{d\theta_{\nu_1}}\frac{d\out_{i_2;\delta_2}}{d\theta_{\nu_2}}\, ,\label{eq:dNTKdef}\\
\mathrm{dd}_{\mathrm{I}}H_{i_0i_1i_2i_3;\delta_0\delta_1\delta_2\delta_3}\equiv& \sum_{\substack{\mu_1,\nu_1\\ \mu_2,\nu_2\\ \mu_3,\nu_3}}\lambda_{\mu_1\nu_1}\lambda_{\mu_2\nu_2}\lambda_{\mu_3\nu_3}\frac{d^3\out_{i_0;\delta_0}}{d\theta_{\mu_1}d\theta_{\mu_2}d\theta_{\mu_3}}\frac{d\out_{i_1;\delta_1}}{d\theta_{\nu_1}}\frac{d\out_{i_2;\delta_2}}{d\theta_{\nu_2}}\frac{d\out_{i_3;\delta_3}}{d\theta_{\nu_3}}\, .\label{eq:ddINTKdef}
\end{align}
They are the neural tangent kernel (NTK), the first-order differential of the NTK (dNTK), and the second-order differential of the NTK (ddNTK) \textit{of type I}, respectively.\footnote{Note that the transformation~\eqref{eq:gauge} leaves $\eta H$, $\eta^2 \mathrm{d}H$, and $\eta^3 \mathrm{dd}_{\mathrm{I}}H$ invariant.}

For the system of equations to be closed, we need to further track the evolution of these dynamical observables: e.g., by Taylor-expanding the update for the NTK~\eqref{eq:NTKdef} in the update for the model parameters~\eqref{eq:Schrodinger}, we get
\begin{align}\label{eq:NTK-update}
&H_{i_0i_1;\delta_0\delta_1}(t+1)\, \\
=&H_{i_0i_1;\delta_0\delta_1}(t)-\eta\sum_{j=1}^{n_L}\sum_{\tra\in\A} \le[\mathrm{d}H_{i_0i_1j;\delta_0\delta_1\tra}(t)+\mathrm{d}H_{i_1i_0j;\delta_1\delta_0\tra}(t)\ri]\frac{d \mathcal{L}_{\A}}{d \out_{j;\tra}}(t)\, \notag\\
&+\frac{1}{2}\eta^2\sum_{j_2,j_3=1}^{n_L}\sum_{\tra_2,\tra_3\in\A}\le[\mathrm{dd}_{\mathrm{I}}H_{i_0i_1j_2j_3;\delta_0\delta_1\tra_2\tra_3}(t)+\mathrm{dd}_{\mathrm{I}}H_{i_1i_0j_2j_3;\delta_1\delta_0\tra_2\tra_3}(t)\ri]\frac{d \mathcal{L}_{\A}}{d \out_{j_2;\tra_2}}(t)\frac{d \mathcal{L}_{\A}}{d \out_{j_3;\tra_3}}(t)\, \notag\\
&+\eta^2\sum_{j_2,j_3=1}^{n_L}\sum_{\tra_2,\tra_3\in\A} \mathrm{dd}_{\mathrm{II}}H_{i_0i_1j_2j_3;\delta_0\delta_1\tra_2\tra_3}(t)\frac{d \mathcal{L}_{\A}}{d \out_{j_2;\tra_2}}(t)\frac{d \mathcal{L}_{\A}}{d \out_{j_3;\tra_3}}(t)+\ldots\, ,\notag
\end{align}
where we additionally introduced the \textit{type-II} ddNTK,
\be
\mathrm{dd}_{\mathrm{II}}H_{i_0i_1i_2i_3;\delta_0\delta_1\delta_2\delta_3}\equiv\sum_{\substack{\mu_1,\nu_1\\ \mu_2,\nu_2\\ \mu_3,\nu_3}}\lambda_{\mu_1\nu_1}\lambda_{\mu_2\nu_2}\lambda_{\mu_3\nu_3}\frac{d^2\out_{i_0;\delta_0}}{d\theta_{\mu_1}d\theta_{\mu_2}}\frac{d^2\out_{i_1;\delta_1}}{d\theta_{\nu_1}d\theta_{\mu_3}}\frac{d\out_{i_2;\delta_2}}{d\theta_{\nu_2}}\frac{d\out_{i_3;\delta_3}}{d\theta_{\nu_3}}\, .\label{eq:ddIINTKdef}
\ee
Here, we see that the dNTK governs the leading-order change in the NTK after a gradient-descent step -- hence its naming -- and, carrying out the same exercise for the dNTK, we would find that a linear combination of the ddNTKs governs the leading-order change in the dNTK. 

In general, these dynamical equations proliferate \textit{ad infinitum}, both because there are an infinite number of Taylor terms for each dynamical equation and because there are an infinite number of dynamical objects with their own update equations.

Thus, at the first sight, these coupled equations look infinitely formidable and intractable. Fortunately, in some regimes -- specifically, when the scale $\gamma=L/n^{1-s}$~\eqref{eq:main} is small -- we can systematically truncate the Taylor expansions and the resulting system of coupled dynamical equations can be solved through perturbation theory~\cite{dyer2019asymptotics,PDLT}. What's more, for a given optimizer, the solution at the end of training can be expressed solely in terms of the dynamical objects \textit{at initialization}. Therefore, we can mostly focus our study on the statistics and scalings of these dynamical objects at initialization, from which we can read off their statistics and scalings after training.\footnote{For this Taylor approach to work, the activation function needs to be smooth. This excludes $\texttt{ReLU}$ activation function from our consideration due to its kink at origin: see Ref.~\cite{PDLT} for more details on this subtlety. That said, empirically there is no observed superiority of \texttt{ReLU} over its smooth variants like \texttt{GELU} and \texttt{SWISH} -- see, e.g., Ref.~\cite{liu2022convnet} -- and hence this theoretical limitation is not practically limiting.}

With those remarks in mind, in the next section, we'll study the network output and NTK at initialization while, in the section after that, we'll study the dNTK and ddNTKs at initialization. Meanwhile, higher-order differentials of the NTK are consigned to Appendix~\ref{app:hierarchy}.

\section{Deriving a Meta-Principled Family}\label{sec:principles}
In this section, we'll use the principle of criticality for hidden-layer preactivations (\S\ref{subsec:criticality}) and the principle of learning-rate equivalence for neural tangent kernels (\S\ref{subsec:equivalence}) to narrow the general class of $p_{\ell}q_{\ell}r$ hyperparameter scaling strategies down to a one-parameter family (\S\ref{subsec:family}).

\subsection{The Principle of Criticality at Meta}\label{subsec:criticality}
As a given input signal propagates forward through many layers of a deep neural network,  the signal gets hit by many similar mathematical operations one after another. In turn, we generically expect that the forward signal either exponentially explodes or exponentially vanishes in magnitude, both of which would inhibit learning. The solution to this problem was proposed in a series of papers~\cite{poole2016exponential,raghu2017expressive,schoenholz2016deep} and later extended in Ref.~\cite{PDLT}. In particular, for the NT scaling with $p_{\ell}=0$, the resulting \textit{principle of criticality} gives a general prescription for tuning initialization hyperparameters $C_b^{(\ell)}$ and $C_W^{(\ell)}$ so as to ensure that signals stay of the same order for all layers.

Here, we'll apply this principle of criticality at a coarser -- or more \textit{meta} -- level, requiring that forward signals stay \textit{parametrically} of order one in all the hidden layers at initialization. Without such requirement, we would have a severe numerical problem for deep neural networks, with the values of preactivations differing by orders of magnitude from layer to layer.\footnote{Yet another problem would be that, as we scale up models in width, the hidden-layer preactivations would explore different ranges of the activation function from scale to scale and hence perceive it as having different characteristics (with a possible exception of scale-invariant activation functions such as \texttt{linear} and \texttt{ReLU} activation functions).}

Specifically, at the point of initialization, the first-layer preactivations~\eqref{eq:MLP1st} have the covariance
\begin{align}
\E{z^{(1)}_{i_1;\delta_1}z^{(1)}_{i_2;\delta_2}}=\frac{1}{n^{p_1}}\le[C_b^{(1)}+C_W^{(1)}\le(\frac{1}{n_0}\sum_{j=1}^{n_0}x_{j;\delta_1}x_{j;\delta_2}\ri)\ri]\delta_{i_1i_2}\, ,
\end{align}
where we used our general initialization hyperparameter scaling strategies~\eqref{eq:init}. Similarly, the $(\ell+1)$-th-layer preactivations~\eqref{eq:MLPrec} have the covariance
\begin{align}
\E{z^{(\ell+1)}_{i_1;\delta_1}z^{(\ell+1)}_{i_2;\delta_2}}=\frac{1}{n^{p_{\ell+1}}}\le\{C_b^{(\ell+1)}+C_W^{(\ell+1)}\frac{1}{n_{\ell}}\sum_{j=1}^{n_{\ell}}\E{\sigma\le(z^{(\ell)}_{j;\delta_1}\ri)\sigma\le(z^{(\ell)}_{j;\delta_2}\ri)}\ri\}\delta_{i_1i_2}\, .
\end{align}
Looking at these covariances, the \textit{meta}-principle of criticality demands that we recursively set
\be\label{eq:preactivation-meta1}
p_1=p_2=\cdots=p_{L-1}=0\, ,
\ee
so that all the hidden-layer preactivations -- and activations -- stay of order one.\footnote{Surely, it appears intuitively obvious that the expectation value $\E{\sigma\le(z^{(\ell)}_{j;\delta_1}\ri)\sigma\le(z^{(\ell)}_{j;\delta_2}\ri)}$ is of order one for a generic activation function $\sigma(z)$ when the $\ell$-th-layer preactivations are of order one. To be surer, for sufficiently large $n$, this expectation value can be turned into a series of Gaussian integrals through the $1/n$ expansion and evaluated order by order~\cite{PDLT}.}

Extremely importantly, differently from the prior work~\cite{poole2016exponential,raghu2017expressive,schoenholz2016deep,PDLT}, we intentionally left out the output layer at $\ell=L$ from this
consideration. From our perspective, the ingenuity of Ref.~\cite{yang2021tensor} lies in realizing that we don't have to apply the meta-principle of criticality to the output layer, which leaves one free parameter,
\be\label{eq:preactivation-meta2}
p_{L}\equiv s\, .
\ee
Specifically, while a positive value of $s$ results in parametrically small network outputs at initialization, it still allows nontrivial learning as long as the product of the global learning rate and the NTK, $\eta H$, is of order one so that the network outputs acquire order-one values upon evolving with the update equation~\eqref{eq:output-update}. We'll come back to this point in \S\ref{subsec:family}.

\subsection{The Principle of Learning-Rate Equivalence at Meta}\label{subsec:equivalence}
Similarly to forward signals, we also need to ensure that backward signals don't exponentially explode or exponentially vanish. More precisely, the gradients of the network outputs with respect to the model parameters must stay of the same order from layer to layer, so that we can stably train networks with gradient-based optimization algorithms.

A convenient way to diagnose this backward signal propagation is to look at the NTK~\eqref{eq:NTKdef},
\be
\NTK_{i_0i_1;\delta_0\delta_1}\equiv\sum_{\mu,\nu} \lambda_{\mu\nu}\frac{d\out_{i_0;\delta_0}}{d\theta_{\mu}}\frac{d\out_{i_1;\delta_1}}{d\theta_{\nu}}\, ,\label{eq:NTKdef2}
\ee
which essentially is a square of the gradients. Here, we've decided to put a hat on the NTK to emphasize that we'll be studying its scaling at the point of initialization.\footnote{We note that in the so-called NTK limit~\cite{jacot2018neural} -- i.e., in the infinite-width limit with the NT scaling and with fixed depth -- the NTK is frozen at its value at initialization, so this hatting is redundant. But at any finite width -- and also in some proper infinite-model-size limits -- the NTK evolves due to nonzero dNTK and ddNTKs.}

To assess its scaling and statistics, it is further convenient to introduce the $\ell$-th-layer NTK~\cite{jacot2018neural,PDLT},
\be
\NTK_{i_0i_1;\delta_0\delta_1}^{(\ell)}\equiv\sum_{\mu,\nu} \lambda_{\mu\nu}\frac{dz^{(\ell)}_{i_0;\delta_0}}{d\theta_{\mu}}\frac{dz^{(\ell)}_{i_1;\delta_1}}{d\theta_{\nu}}\, ,\label{eq:NTKdefell}\\
\ee
which reproduces the output NTK~\eqref{eq:NTKdef2}  when $\ell=L$. More importantly, it obeys the forward equation
\begin{align}\label{eq:NTKforward}
\NTK^{(\ell+1)}_{i_1i_2;\delta_1\delta_2}=&\frac{1}{n^{q_{\ell+1}}}\le\{\Lb^{(\ell+1)}+\LW^{(\ell+1)}\le[\frac{1}{n_{\ell}}\sum_{j=1}^{n_{\ell}}\sigma\le(z^{(\ell)}_{j;\delta_1}\ri)\sigma\le(z^{(\ell)}_{j;\delta_2}\ri)\ri]\ri\}\delta_{i_1i_2}\, \\
&+\sum_{j_1,j_2=1}^{n_{\ell}}W^{(\ell+1)}_{i_1j_1}W^{(\ell+1)}_{i_2j_2}\sigma'\le(z^{(\ell)}_{j_1;\delta_1}\ri)\sigma'\le(z^{(\ell)}_{j_2;\delta_2}\ri)\NTK^{(\ell)}_{j_1j_2;\delta_1\delta_2}\, ,\notag
\end{align}
which can be derived by plugging in our general training hyperparameter scaling strategies~\eqref{eq:train} and using the chain rule with the familiar factor
\be\label{eq:chain}
\frac{dz_{i;\delta}^{(\ell+1)}}{dz_{j;\delta}^{(\ell)}}=W^{(\ell+1)}_{ij}\sigma'\le(z^{(\ell)}_{j;\delta}\ri)\, ,
\ee
that also appears in the backpropagation algorithm. The first term on the right-hand side of the NTK forward equation~\eqref{eq:NTKforward} captures the additive contributions from the gradients with respect to the $(\ell+1)$-th-layer model parameters; the second term captures the cumulative contributions from the previous layers, with the $\ell$-th-layer NTK multiplied by two backpropagation factors -- though we are emphatically going forward here.

Now, the principle of learning-rate equivalence~\cite{PDLT} gives a prescription for fine-tuning the training hyperparameters $\lambda_b^{(\ell)}$ and $\lambda_W^{(\ell)}$ so as to ensure that gradients from all layers contribute equally to the output NTK -- preconditioned on the principle of criticality that ensures hidden-layer preactivations neither exponentially explode nor exponentially vanish. Here, we apply the same principle -- again at a more \textit{meta} level -- requiring that the two terms on the right-hand side of the NTK forward equation~\eqref{eq:NTKforward} are \textit{parametrically} of the same order.

Putting that in action, on the one hand, the first term is of order $O\le(1/n^{q_{\ell+1}}\ri)$.\footnote{Here and right after, we are preconditioning our scaling analysis on the meta-principle of criticality, i.e., that hidden-layer preactivations -- and in turn hidden-layer activations and their derivatives --  are parametrically of order one.} On the other hand, the second term is of order $O\le(1/n^{p_{\ell+1}}\ri)\times O\big(\NTK^{(\ell)}\big)$.\footnote{To be a little more verbose, the factor of $1/n_{\ell}$ coming from the base NT scaling for the initial weights cancels out with the summation over $j=1,\ldots,n_{\ell}$. To get the flavor of it, we can take the expectation value of the NTK forward equation~\eqref{eq:NTKforward}, which, together with the initialization distributions~\eqref{eq:init}, yields
\begin{align}
\E{\NTK^{(\ell+1)}_{i_1i_2;\delta_1\delta_2}}=&\frac{1}{n^{q_{\ell+1}}}\le\{\Lb^{(\ell+1)}+\LW^{(\ell+1)}\E{\frac{1}{n_{\ell}}\sum_{j=1}^{n_{\ell}}\sigma\le(z^{(\ell)}_{j;\delta_1}\ri)\sigma\le(z^{(\ell)}_{j;\delta_2}\ri)}\ri\}\delta_{i_1i_2}\, \\
&+\frac{1}{n^{p_{\ell+1}}}\le\{C_W^{(\ell+1)}\E{\frac{1}{n_{\ell}}\sum_{j=1}^{n_{\ell}}\sigma'\le(z^{(\ell)}_{j;\delta_1}\ri)\sigma'\le(z^{(\ell)}_{j;\delta_2}\ri)\NTK^{(\ell)}_{jj;\delta_1\delta_2}}\ri\}\delta_{i_1i_2}\, ,\notag
\end{align}
wherein we see the nice pairings of $1/n_{\ell}$ and $\sum_{j=1}^{n_{\ell}}$. For an even more verbose analysis, including the effects of the NTK fluctuations, please cf.~Ref.~\cite{PDLT}. In particular, we can show that the contributions from the NTK fluctuations are  suppressed by a factor of $O\le(L/n\ri)$.}
To finish balancing these two terms, noting that the first-layer NTK
\begin{align}\label{eq:firstNTK}
\NTK^{(1)}_{i_1i_2;\delta_1\delta_2}=\frac{1}{n^{q_1}}\le[\Lb^{(1)}+\LW^{(1)}\le(\frac{1}{n_0}\sum_{j=1}^{n_0}x_{j;\delta_1}x_{j;\delta_2}\ri)\ri]\delta_{i_1i_2}\, 
\end{align}
scales as $O\le(1/n^{q_{1}}\ri)$, hence consistently and inductively assuming that the $\ell$-th-layer NTK scales as $O\le(1/n^{q_{\ell}}\ri)$, and recalling the scalings of the two terms described in the two previous sentences, we get by induction
\be\label{eq:NTK-meta-master}
q_{\ell+1}=p_{\ell+1}+q_{\ell}\,  \ \ \ \text{for}\ \ \ \ell=1,\ldots, L-1\, .
\ee

Therefore, recalling that $p_{1}=\cdots=p_{L-1}=0$~\eqref{eq:preactivation-meta1} and $p_{L}=s$~\eqref{eq:preactivation-meta2},
the meta-principle of learning-rate equivalence further demands that we recursively set
\be\label{eq:NTK-meta1}
q_{1}=q_{2}=\cdots q_{L-1}\equiv t\, 
\ee
and
\be\label{eq:NTK-meta2}
q_{L}=t+s\, .
\ee

\subsection{Meta-Principled Family}\label{subsec:family}
At this point, unlike we promised, it appears as if we still got three parameters at our disposal: $r$ from the global learning rate $\eta$~\eqref{eq:eta-scale}, $s$ from the last-layer preactivation scalings~\eqref{eq:preactivation-meta2}, and $t$ from the bulk of the learning-rate tensor~\eqref{eq:NTK-meta1}. Let's quickly cut these three down to one.

First, referring back to the network evolution~\eqref{eq:output-update}, to have some learning in finite steps, we need the product of learning rate $\eta=O\le(n^{r}\ri)$ and the output NTK $\NTK=O\le(1/n^{q_{L}}\ri)$ to be of order one. This requires that
\be\label{eq:nontrivial-learning}
r=q_{L}=t+s\, ,
\ee
reducing the three down to two.\footnote{In general, there is also a nontrivial depth scaling coming from $\NTK^{(L)}$. Here we account for that through $\etanorm\lambda_{b,W}^{(\ell)}$.}

Next, to cut the remaining two down to one, we recall that there is trivial gauge redundancy~\eqref{eq:gauge} without any physical consequence, which can be used to gauge away one redundant exponent. Specifically, choosing the transformation~\eqref{eq:gauge} with $g=-t$, we finally get
\be\label{eq:ultimate-exponents}
p_{1}=\cdots=p_{L-1}=0\, , \ \ \ p_{L}=s\, , \ \ \ q_{1}=\cdots=q_{L-1}=0\, , \ \ \ q_{L}=s\, , \ \ \ r=s\, .
\ee

In summary, we are left with only one \textit{meta}parameter, $s$, that parametrizes the meta-principled family of hyperparameter scaling strategies:
\begin{align}
\E{b_{i_1}^{(\ell)}b_{i_2}^{(\ell)}}=C_b^{(\ell)}\delta_{i_1i_2}\, , \ \ \ \E{W_{i_1j_1}^{(\ell)}W_{i_2j_2}^{(\ell)}}=&\frac{C_W^{(\ell)}}{n_{\ell-1}}\delta_{i_1i_2}\delta_{j_1j_2}\, ,\ \ \ \text{for}\ \ \ \ell=1,\ldots,L-1\, ,\label{eq:meta-scaling1}\\
\E{b_{i_1}^{(L)}b_{i_2}^{(L)}}=\frac{C_b^{(L)}}{n_{L-1}^{s}}\delta_{i_1i_2}\, , \ \ \ \E{W_{i_1j_1}^{(L)}W_{i_2j_2}^{(L)}}=&\frac{C_W^{(L)}}{n_{L-1}^{1+s}}\delta_{i_1i_2}\delta_{j_1j_2}\, ,\label{eq:meta-scaling2}\\
\lambda_{b_{i_1}^{(\ell)}b_{i_2}^{(\ell)}}=\lambda_b^{(\ell)}\delta_{i_1i_2}\, , \ \ \ \lambda_{W_{i_1j_1}^{(\ell)}W_{i_2j_2}^{(\ell)}}=&\frac{\lambda_W^{(\ell)}}{n_{\ell-1}}\delta_{i_1i_2}\delta_{j_1j_2}\, ,\ \ \ \text{for}\ \ \ \ell=1,\ldots,L-1\, ,\label{eq:meta-scaling3}\\
\lambda_{b_{i_1}^{(L)}b_{i_2}^{(L)}}=\frac{1}{n_{L-1}^{s}}\le(\lambda_b^{(L)}\delta_{i_1i_2}\ri)\, , \ \ \ \lambda_{W_{i_1j_1}^{(L)}W_{i_2j_2}^{(L)}}=&\frac{\lambda_W^{(L)}}{n_{L-1}^{1+s}}\delta_{i_1i_2}\delta_{j_1j_2}\, ,\label{eq:meta-scaling4}\\
\eta=&n^{s}_{L-1}\etanorm\, ,\label{eq:meta-scaling5}
\end{align}
where, for concreteness, we set $n\equiv n_{L-1}$. In words, with respect to the NT scaling, we suppress the initialization hyperparameters and the learning-rate tensor components by a factor of $1/n^s$ for the last-layer biases and weights, while encouraging the global learning rate by a factor of  $n^s$.\footnote{To apply the same meta-principles to more general architectures, the pragmatic rule of thumb is to just replace $n_{\ell-1}$ by the number of incoming signal components, often denoted as $\fanin$ in PyTorch documentations. For hidden layers, our scaling strategies for initialization hyperparameters are then generally in line with the Xavier~\cite{glorot2010understanding} and Kaiming~\cite{he2015delving} initialization schemes.}

This family interpolates the NT scaling~\cite{jacot2018neural} at $s=0$ and the MU scaling~\cite{yang2021tensor} at $s=1$. In fact, these two strategies lie at the two ends of the sensible interval for the metaparameter $s\in[0,1]$: $s\geq0$ because, otherwise, the network outputs would be parametrically large; $s\leq1$ because, otherwise, the differentials of the NTK would be parametrically larger, as we'll see next.\footnote{At this point, for those familiar with Ref.~\cite{yang2021tensor}, it may be helpful to pause and carry out an in-depth comparison between this part of our results against theirs.\\
Here are similarities. In Ref.~\cite{yang2021tensor}, they invoked the ``stability condition'' to put hard constraints, $p_{1}=\cdots p_{L-1}=0$ (along with various soft constraints, i.e., inequalities), much like our meta-principle of criticality did. They then invoked the ``nontriviality condition'' to put a hard constraint $r=q_L$ -- or another more complicated condition that is also satisfied by our family -- much like our condition of finite learning~\eqref{eq:nontrivial-learning}, $\eta \NTK=O\le(1\ri)$, did. They further noticed the same symmetry under the transformation~\eqref{eq:gauge}.  These together cut the $2L+1$ exponents down to $L$, so far in the same way as ours.\\
Here are differences. They invoked the ``feature learning condition,'' $0=1+\min_{\ell=1}^{L-1}(q_{\ell}-r)+\min\le(q_L-r, \frac{p_{L}+1}{2}\ri)$. This condition was put forth to preserve feature learning in the infinite-width limit \textit{with fixed depth}. To single out the MU scaling, they actually provoked stronger conditions, the ``maximal-update conditions,'' $0=1+(q_{\ell}-r)+\min\le(q_L-r, \frac{p_{L}+1}{2}\ri)$ for $\ell=1,\ldots,L-1$ (which automatically satisfy the ``feature learning condition'') and the ``maximal-initialization condition,'' $0=\frac{p_L+1}{2}+\min_{\ell=1}^{L-1}(q_{\ell}-r)+\min\le(q_L-r, \frac{p_{L}+1}{2}\ri)$. Our family -- which instead arose from the meta-principle of learning-rate equivalence -- connects this MU scaling at $s=1$ all the way down to the NT scaling at $s=0$ while preserving representation learning for $s<1$ by \textit{scaling up the depth} with the width.}

\section{Computing the Degree of Representation Learning}\label{sec:seeing}
In this section, we'll analyze the scalings of the dNTK~\eqref{eq:dNTKdef} in \S\ref{subsec:dNTK} and ddNTKs~\eqref{eq:ddINTKdef} \eqref{eq:ddIINTKdef} in \S\ref{subsec:ddNTKs}. To motivate these calculations, let's look back at the NTK update equation~\eqref{eq:NTK-update}. As noted there, these differentials control changes in the NTK during training. Viewing the kernel as a sum of features that comprise representation, the magnitudes of these differentials in turn quantify the degree of leading-order representation learning~\cite{PDLT}, which we'll compute here for wide and deep neural networks. This computation results in the emergent scale $\gamma$~\eqref{eq:meta-up}.

\subsection{dNTK}\label{subsec:dNTK}
Following our recursive analysis of the NTK, let's define the $\ell$-th-layer dNTK as
\be
\dNTK_{i_0i_1i_2;\delta_0\delta_1\delta_2}^{(\ell)}\equiv\sum_{\substack{\mu_1,\nu_1\\ \mu_2,\nu_2}}\lambda_{\mu_1\nu_1}\lambda_{\mu_2\nu_2}\frac{d^2z^{(\ell)}_{i_0;\delta_0}}{d\theta_{\mu_1}d\theta_{\mu_2}}\frac{dz^{(\ell)}_{i_1;\delta_1}}{d\theta_{\nu_1}}\frac{dz^{(\ell)}_{i_2;\delta_2}}{d\theta_{\nu_2}}\, ,\\
\ee
where we've again hatted it to emphasize that we are analyzing its scaling at initialization. This object reproduces the output dNTK~\eqref{eq:dNTKdef} when $\ell=L$ while it vanishes in the first layer $\ell=1$ because the first-layer preactivations~\eqref{eq:MLP1st} are linear in the model parameters. In the middle -- nearly reprinting an equation from Ref.~\cite{PDLT} -- it obeys the following forward equation:
\begin{align}\label{eq:dNTKforward}
\dNTK_{i_0i_1i_2;\delta_0\delta_1\delta_2}^{(\ell+1)}&=\frac{1}{n^{q_{\ell+1}}}\frac{\lambda_W^{(\ell+1)}}{n_{\ell}}\delta_{i_0i_1}\sum_{j_0,j_2=1}^{n_{\ell}}W^{(\ell+1)}_{i_2j_2}\sigma'\le(z^{(\ell)}_{j_0;\delta_0}\ri)\sigma\le(z^{(\ell)}_{j_0;\delta_1}\ri)\sigma'\le(z^{(\ell)}_{j_2;\delta_2}\ri)\NTK_{j_0j_2;\delta_0\delta_2}^{(\ell)}\, \\
&+\frac{1}{n^{q_{\ell+1}}}\frac{\lambda_W^{(\ell+1)}}{n_{\ell}}\delta_{i_0i_2}\sum_{j_0,j_1=1}^{n_{\ell}}W^{(\ell+1)}_{i_1j_1}\sigma'\le(z^{(\ell)}_{j_0;\delta_0}\ri)\sigma'\le(z^{(\ell)}_{j_1;\delta_1}\ri)\sigma\le(z^{(\ell)}_{j_0;\delta_2}\ri)\NTK_{j_0j_1;\delta_0\delta_1}^{(\ell)}\, \notag\\
&+\sum_{j_0,j_1,j_2=1}^{n_{\ell}}W^{(\ell+1)}_{i_0j_0}W^{(\ell+1)}_{i_1j_1}W^{(\ell+1)}_{i_2j_2}\sigma'\le(z^{(\ell)}_{j_0;\delta_0}\ri)\sigma'\le(z^{(\ell)}_{j_1;\delta_1}\ri)\sigma'\le(z^{(\ell)}_{j_2;\delta_2}\ri) \dNTK_{j_0j_1j_2;\delta_0\delta_1\delta_2}^{(\ell)}\, \notag\\
&+\sum_{j_0,j_1,j_2=1}^{n_{\ell}}W^{(\ell+1)}_{i_0j_0}W^{(\ell+1)}_{i_1j_1}W^{(\ell+1)}_{i_2j_2}\sigma''\le(z^{(\ell)}_{j_0;\delta_0}\ri)\sigma'\le(z^{(\ell)}_{j_1;\delta_1}\ri)\sigma'\le(z^{(\ell)}_{j_2;\delta_2}\ri) \NTK_{j_0j_1;\delta_0\delta_1}^{(\ell)}\NTK_{j_0j_2;\delta_0\delta_2}^{(\ell)}\, .\notag
\end{align}
This can again be derived -- mindlessly object-wise but mindfully index-wise -- by simply plugging in our general training hyperparameter scaling strategies~\eqref{eq:train} and repeatedly using the chain rule.

For the meta-principled family~\eqref{eq:ultimate-exponents}, because $p_{1}=\cdots=p_{L-1}=q_{1}=\cdots=q_{L-1}=0$, the bulk of the dNTK forward equation~\eqref{eq:dNTKforward} stays the same with the NT scaling strategy -- for which the width and depth scalings of the dNTK have already been analyzed in detail~\cite{PDLT}. We thus only need to pay special attention to the equation when $\ell+1=L$. There, for the first two terms on the right-hand side of the forward equation~\eqref{eq:dNTKforward}, with respect to the NT scaling, there is an additional factor of $1/n^s$ coming from $1/n^{q_{L}}$ in front and another additional factor of $1/n^{\frac{s}{2}}$ coming from the last-layer weight; for the last two terms, similarly, there is an additional factor of  $1/n^{\frac{3s}{2}}$ coming from the three last-layer weights. Thus, the gauge-invariant combination, $\eta^2 \dNTK_{i_0i_1i_2;\delta_0\delta_1\delta_2}^{(L)}$, naively comes with an additional factor of $n^{2s}/n^{\frac{3s}{2}}=n^{\frac{s}{2}}$ with respect to the NT scaling. One twist is that it has an odd number of the last-layer weights, so its expectation vanishes; the first nontrivial expectation value is of the form $\E{z_{i;\delta}\eta^2 \dNTK_{i_0i_1i_2;\delta_0\delta_1\delta_2}}$, for which the $s$ dependence cancels out.

All in all, downloading the scaling of this expectation value from Ref.~\cite{PDLT}, we get
\be\label{eq:dNTK-scaling}
\E{z_{i;\delta}^{(L)}\eta^2 \dNTK_{i_0i_1i_2;\delta_0\delta_1\delta_2}^{(L)}}=O\le(\frac{L}{n}\ri)\, .
\ee
Thus, for a general value of $s$, the leading effect of the dNTK at initialization scales like the depth-to-width aspect ratio of the network.\footnote{To be precise, the equation~\eqref{eq:dNTK-scaling} is imprecise. What actually happens in Ref.~\cite{PDLT} on this topic is that \textit{(i)} we study behaviors of signals in deep multilayer perceptrons and find various \textit{universality classes} of activation functions; \textit{(ii)} for each universality class, we solve the forward equations for the preactivations and the NTK, and find and choose the critical initialization hyperparameters $C_{b,W}^{(\ell)}$ and training hyperparameters $\lambda_{b,W}^{(\ell)}$ according to the principles of criticality and learning-rate equivalence so as to maximally avoid the exploding and vanishing problems; \textit{(iii)} we then further solve the forward equations for the dNTK and ddNTKs to read off their width and depth scaling; and \textit{(iv)} we look at dimensionless quantities such as $\E{z\eta^2\dNTK}/\le(\E{\eta\NTK}\ri)^2$ and $\E{z^2}\E{\eta^3\ddNTKI}/\le(\E{\eta\NTK}\ri)^2$,  and -- for the NT scaling strategy -- find $L/n$ scalings for all the leading corrections to the infinite-width limit.\\
For those who could actually parse the last paragraph, there is one additional comment:
in the last step \textit{(iv)}, for the dimensionless combination for the ddNTK contribution, one might worry about the cancellation of the $s$ dependence due to the factor of $\E{z^2}$, which scales like $1/n^s$; however, there is another dimensionless combination with $\E{z^2}$ replaced by $y^2$ with the training labels $y_{i;\tra}$, and this combination does indeed appear in the end-of-training-prediction formula~\cite{PDLT} and contributes dominantly without the $s$ cancellation.}

\subsection{ddNTKs}\label{subsec:ddNTKs}
At this point, all the calculational procedures are familiar, so we'll briefly describe the calculations for the type-I ddNTK in the main text and the ones for the type-II ddNTK in a footnote.

We define the $\ell$-th-layer type-I ddNTK as
\be
\ddNTKI_{i_0i_1i_2i_3;\delta_0\delta_1\delta_2\delta_3}^{(\ell)}\equiv \sum_{\substack{\mu_1,\nu_1\\ \mu_2,\nu_2\\ \mu_3,\nu_3}}\lambda_{\mu_1\nu_1}\lambda_{\mu_2\nu_2}\lambda_{\mu_3\nu_3}\frac{d^3z^{(\ell)}_{i_0;\delta_0}}{d\theta_{\mu_1}d\theta_{\mu_2}d\theta_{\mu_3}}\frac{dz^{(\ell)}_{i_1;\delta_1}}{d\theta_{\nu_1}}\frac{dz^{(\ell)}_{i_2;\delta_2}}{d\theta_{\nu_2}}\frac{dz^{(\ell)}_{i_3;\delta_3}}{d\theta_{\nu_3}}\, ,
\ee
which obeys the following forward equation that can be derived by plugging in our diagonal learning-rate tensor~\eqref{eq:train} and using the chain rule:
\begin{align}\label{eq:ddNTKIforward}
&\ddNTKI_{i_0i_1i_2i_3;\delta_0\delta_1\delta_2\delta_3}^{(\ell+1)}\, \\
&=\frac{1}{n^{q_{\ell+1}}}\frac{\lambda_W^{(\ell+1)}}{n_{\ell}}\delta_{i_0i_1}\sum_{j_0,j_1,j_2,j_3=1}^{n_{\ell}}\delta_{j_0j_1}W^{(\ell+1)}_{i_2j_2}W^{(\ell+1)}_{i_3j_3}\sigma\le(z^{(\ell)}_{j_1;\delta_1}\ri)\sigma'\le(z^{(\ell)}_{j_2;\delta_2}\ri)\sigma'\le(z^{(\ell)}_{j_3;\delta_3}\ri)\, \notag\\
&\ \ \ \ \ \ \ \ \ \ \ \ \ \times\le[\sigma''\le(z^{(\ell)}_{j_0;\delta_0}\ri)\NTK_{j_0j_2;\delta_0\delta_2}^{(\ell)}\NTK_{j_0j_3;\delta_0\delta_3}^{(\ell)}+\sigma'\le(z^{(\ell)}_{j_0;\delta_0}\ri)\dNTK_{j_0j_2j_3;\delta_0\delta_2\delta_3}^{(\ell)}\ri]\, \notag\\
&+\frac{1}{n^{q_{\ell+1}}}\frac{\lambda_W^{(\ell+1)}}{n_{\ell}}\delta_{i_0i_2}\sum_{j_0,j_1,j_2,j_3=1}^{n_{\ell}}\delta_{j_0j_2}W^{(\ell+1)}_{i_3j_3}W^{(\ell+1)}_{i_1j_1}\sigma\le(z^{(\ell)}_{j_2;\delta_2}\ri)\sigma'\le(z^{(\ell)}_{j_3;\delta_3}\ri)\sigma'\le(z^{(\ell)}_{j_1;\delta_1}\ri)\, \notag\\
&\ \ \ \ \ \ \ \ \ \ \ \ \ \times\le[\sigma''\le(z^{(\ell)}_{j_0;\delta_0}\ri)\NTK_{j_0j_3;\delta_0\delta_3}^{(\ell)}\NTK_{j_0j_1;\delta_0\delta_1}^{(\ell)}+\sigma'\le(z^{(\ell)}_{j_0;\delta_0}\ri)\dNTK_{j_0j_3j_1;\delta_0\delta_3\delta_1}^{(\ell)}\ri]\, \notag\\
&+\frac{1}{n^{q_{\ell+1}}}\frac{\lambda_W^{(\ell+1)}}{n_{\ell}}\delta_{i_0i_3}\sum_{j_0,j_1,j_2,j_3=1}^{n_{\ell}}\delta_{j_0j_3}W^{(\ell+1)}_{i_1j_1}W^{(\ell+1)}_{i_2j_2}\sigma\le(z^{(\ell)}_{j_3;\delta_3}\ri)\sigma'\le(z^{(\ell)}_{j_1;\delta_1}\ri)\sigma'\le(z^{(\ell)}_{j_2;\delta_2}\ri)\, \notag\\
&\ \ \ \ \ \ \ \ \ \ \ \ \ \times\le[\sigma''\le(z^{(\ell)}_{j_0;\delta_0}\ri)\NTK_{j_0j_1;\delta_0\delta_1}^{(\ell)}\NTK_{j_0j_2;\delta_0\delta_2}^{(\ell)}+\sigma'\le(z^{(\ell)}_{j_0;\delta_0}\ri)\dNTK_{j_0j_1j_2;\delta_0\delta_1\delta_2}^{(\ell)}\ri]\, \notag\\
&+\sum_{j_0,j_1,j_2,j_3=1}^{n_{\ell}}W^{(\ell+1)}_{i_0j_0}W^{(\ell+1)}_{i_1j_1}W^{(\ell+1)}_{i_2j_2}W^{(\ell+1)}_{i_3j_3}\sigma'\le(z^{(\ell)}_{j_1;\delta_1}\ri)\sigma'\le(z^{(\ell)}_{j_2;\delta_2}\ri)\sigma'\le(z^{(\ell)}_{j_3;\delta_3}\ri)\, \notag\\
&\ \ \ \ \times\Big[\sigma'\le(z^{(\ell)}_{j_0;\delta_0}\ri)\ddNTKI_{j_0j_1j_2j_3;\delta_0\delta_1\delta_2\delta_3}^{(\ell)}+\sigma''\le(z^{(\ell)}_{j_0;\delta_0}\ri)\dNTK_{j_0j_1j_2;\delta_0\delta_1\delta_2}^{(\ell)}\NTK^{(\ell)}_{j_0j_3;\delta_0\delta_3}\, \notag\\
&\ \ \ \ \ \ \ \ +\sigma''\le(z^{(\ell)}_{j_0;\delta_0}\ri)\dNTK_{j_0j_2j_3;\delta_0\delta_2\delta_3}^{(\ell)}\NTK^{(\ell)}_{j_0j_1;\delta_0\delta_1}+\sigma''\le(z^{(\ell)}_{j_0;\delta_0}\ri)\dNTK_{j_0j_3j_1;\delta_0\delta_3\delta_1}^{(\ell)}\NTK^{(\ell)}_{j_0j_2;\delta_0\delta_2}\, \notag\\
&\ \ \ \ \ \ \  \ \ \ \ \ \ \ \ \ \ \ \ \ \ \ \ \ \ \ \ \ \ \ \ \ \ \ \ \ \ \ \ \ \ \ \ \ \ \ \ \ +\sigma'''\le(z^{(\ell)}_{j_0;\delta_0}\ri)\NTK^{(\ell)}_{j_0j_1;\delta_0\delta_1}\NTK^{(\ell)}_{j_0j_2;\delta_0\delta_2}\NTK^{(\ell)}_{j_0j_3;\delta_0\delta_3}\Big]\, .\notag
\end{align}
With respect to the NT scaling,  all the terms get the multiplicative factor of $1/n^{2s}$ at $\ell+1=L$, so, once again using the result from Ref.~\cite{PDLT} for $s=0$, the gauge-invariant combination scales as
\be
\E{\eta^3 \ddNTKI^{(L)}_{i_0i_1i_2i_3;\delta_0\delta_1\delta_2\delta_3}}=O\le(\frac{L}{n^{1-s}}\ri)\, 
\ee
for wide and deep neural networks.\footnote{As for  the type-II ddNTK, we define the $\ell$-th-layer version as
\be
\ddNTKII_{i_0i_1i_2i_3;\delta_0\delta_1\delta_2\delta_3}^{(\ell)}\equiv\sum_{\substack{\mu_1,\nu_1\\ \mu_2,\nu_2\\ \mu_3,\nu_3}}\lambda_{\mu_1\nu_1}\lambda_{\mu_2\nu_2}\lambda_{\mu_3\nu_3}\frac{d^2z^{(\ell)}_{i_0;\delta_0}}{d\theta_{\mu_1}d\theta_{\mu_2}}\frac{d^2z^{(\ell)}_{i_1;\delta_1}}{d\theta_{\nu_1}d\theta_{\mu_3}}\frac{dz^{(\ell)}_{i_2;\delta_2}}{d\theta_{\nu_2}}\frac{dz^{(\ell)}_{i_3;\delta_3}}{d\theta_{\nu_3}}\, ,
\ee
which obeys the forward equation
\begin{align}\label{eq:ddNTKIIforward}
&\ddNTKII_{i_0i_1i_2i_3;\delta_0\delta_1\delta_2\delta_3}^{(\ell+1)}\, \\
&=\frac{1}{n^{2q_{\ell+1}}}\le(\frac{\lambda_W^{(\ell+1)}}{n_{\ell}}\ri)^2\delta_{i_0i_2}\delta_{i_1i_3}\sum_{j_0,j_1=1}^{n_{\ell}}\sigma'\le(z^{(\ell)}_{j_0;\delta_0}\ri)\sigma'\le(z^{(\ell)}_{j_1;\delta_1}\ri)\sigma\le(z^{(\ell)}_{j_0;\delta_2}\ri)\sigma\le(z^{(\ell)}_{j_1;\delta_3}\ri)\NTK^{(\ell)}_{j_0j_1;\delta_0\delta_1}\, \notag\\
&+\frac{1}{n^{q_{\ell+1}}}\frac{\lambda_W^{(\ell+1)}}{n_{\ell}}\delta_{i_0i_1}\sum_{j_0,j_1,j_2,j_3=1}^{n_{\ell}}\delta_{j_0j_1}W^{(\ell+1)}_{i_2j_2}W^{(\ell+1)}_{i_3j_3}\, \notag\\
&\ \ \ \ \ \ \ \ \ \ \ \ \ \times\sigma'\le(z^{(\ell)}_{j_0;\delta_0}\ri)\sigma'\le(z^{(\ell)}_{j_1;\delta_1}\ri)\sigma'\le(z^{(\ell)}_{j_2;\delta_2}\ri)\sigma'\le(z^{(\ell)}_{j_3;\delta_3}\ri)\NTK^{(\ell)}_{j_0j_2;\delta_0\delta_2}\NTK^{(\ell)}_{j_1j_3;\delta_1\delta_3}\, \notag\\
&+\frac{1}{n^{q_{\ell+1}}}\frac{\lambda_W^{(\ell+1)}}{n_{\ell}}\delta_{i_0i_2}\sum_{j_0,j_1,j_2,j_3=1}^{n_{\ell}}\delta_{j_0j_2}W^{(\ell+1)}_{i_1j_1}W^{(\ell+1)}_{i_3j_3}\sigma\le(z^{(\ell)}_{j_2;\delta_2}\ri)\sigma'\le(z^{(\ell)}_{j_3;\delta_3}\ri)\sigma'\le(z^{(\ell)}_{j_0;\delta_0}\ri)\, \notag\\
&\ \ \ \ \ \ \ \ \ \ \ \ \ \times\le[\sigma''\le(z^{(\ell)}_{j_1;\delta_1}\ri)\NTK_{j_1j_0;\delta_1\delta_0}^{(\ell)}\NTK_{j_1j_3;\delta_1\delta_3}^{(\ell)}+\sigma'\le(z^{(\ell)}_{j_1;\delta_1}\ri)\dNTK_{j_1j_0j_3;\delta_1\delta_0\delta_3}^{(\ell)}\ri]\, \notag\\
&+\frac{1}{n^{q_{\ell+1}}}\frac{\lambda_W^{(\ell+1)}}{n_{\ell}}\delta_{i_1i_3}\sum_{j_0,j_1,j_2,j_3=1}^{n_{\ell}}\delta_{j_1j_3}W^{(\ell+1)}_{i_0j_0}W^{(\ell+1)}_{i_2j_2}\sigma\le(z^{(\ell)}_{j_3;\delta_3}\ri)\sigma'\le(z^{(\ell)}_{j_2;\delta_2}\ri)\sigma'\le(z^{(\ell)}_{j_1;\delta_1}\ri)\, \notag\\
&\ \ \ \ \ \ \ \ \ \ \ \ \ \times\le[\sigma''\le(z^{(\ell)}_{j_0;\delta_0}\ri)\NTK_{j_0j_1;\delta_0\delta_1}^{(\ell)}\NTK_{j_0j_2;\delta_0\delta_2}^{(\ell)}+\sigma'\le(z^{(\ell)}_{j_0;\delta_0}\ri)\dNTK_{j_0j_1j_2;\delta_0\delta_1\delta_2}^{(\ell)}\ri]\, \notag\\
&+\sum_{j_0,j_1,j_2,j_3=1}^{n_{\ell}}W^{(\ell+1)}_{i_0j_0}W^{(\ell+1)}_{i_1j_1}W^{(\ell+1)}_{i_2j_2}W^{(\ell+1)}_{i_3j_3}\sigma'\le(z^{(\ell)}_{j_2;\delta_2}\ri)\sigma'\le(z^{(\ell)}_{j_3;\delta_3}\ri)\, \notag\\
&\ \ \ \ \ \times\Big[\sigma'\le(z^{(\ell)}_{j_0;\delta_0}\ri)\sigma'\le(z^{(\ell)}_{j_1;\delta_1}\ri)\ddNTKII_{j_0j_1j_2j_3;\delta_0\delta_1\delta_2\delta_3}^{(\ell)}+\sigma''\le(z^{(\ell)}_{j_0;\delta_0}\ri)\sigma''\le(z^{(\ell)}_{j_1;\delta_1}\ri)\NTK_{j_0j_1;\delta_0\delta_1}^{(\ell)}\NTK^{(\ell)}_{j_0j_2;\delta_0\delta_2}\NTK^{(\ell)}_{j_1j_3;\delta_1\delta_3}\, \notag\\
&\ \ \ \ \ \ \ \ +\sigma'\le(z^{(\ell)}_{j_0;\delta_0}\ri)\sigma''\le(z^{(\ell)}_{j_1;\delta_1}\ri)\NTK^{(\ell)}_{j_1j_3;\delta_1\delta_3}\dNTK_{j_0j_1j_2;\delta_0\delta_1\delta_2}^{(\ell)}+\sigma'\le(z^{(\ell)}_{j_1;\delta_1}\ri)\sigma''\le(z^{(\ell)}_{j_0;\delta_0}\ri)\NTK^{(\ell)}_{j_0j_2;\delta_0\delta_2}\dNTK_{j_1j_0j_3;\delta_1\delta_0\delta_3}^{(\ell)}\Big]\, ,\notag
\end{align}
which gets the multiplicative factor of $1/n^{2s}$ at $\ell+1=L$ with respect to the NT scaling, and thus
\be
\E{\eta^3 \ddNTKII^{(L)}}=O\le(\frac{L}{n^{1-s}}\ri)\, .
\ee}

Through these calculations, there emerges the scale,
\be\label{eq:meta-up}
\gamma\equiv \frac{L}{n^{1-s}}\, ,
\ee
that controls the effect of the ddNTKs. For $s=0$, this is of the same order as the effect of the dNTK~\eqref{eq:dNTK-scaling}, while for any positive $s>0$ this contribution is more dominant. In general, the emergent scale $\gamma$ characterizes the amount of the leading-order change in the NTK and hence the degree of representation learning for wide and deep neural networks.\footnote{To be a bit more precise, the depth scaling is derived for deep neural networks, and for shallow neural networks like the one-hidden-layer ones with $L=2$, the ``$L$'' in the numerator of the scale $\gamma=L/n^{1-s}$ should just be thought of as an order-one number.}
When small, it also serves as a perturbative parameter that quantifies the leading-order change to the frozen NTK limit and, at least for one-hidden-layer networks, we  explicitly show in Appendix~\ref{app:hierarchy} that the higher-order differentials of the NTK are suppressed by higher-order powers of $\gamma$.

\section{Casting the Web of Effective Theories}\label{sec:engulf}
Thus far, we've derived a family of hyperparameter scaling strategies, meta-parametrized by $s\in[0,1]$, from the meta-principle of criticality and the meta-principle of learning-rate equivalence.\footnote{In a sense, our derivation gives a no-go theorem for other hyperparameter scaling strategies. To hedge, we point out that no-go theorems sometimes contain subtle loopholes in their assumptions which, upon realization and proper relaxation, lead to interesting phenomena: cf.~the relation between the Coleman-Mandula theorem~\cite{coleman1967all} and supersymmetry~\cite{weinberg2005quantum}.} We've then assessed the scalings of representation learning as a function of the hidden-layer width $n$ and the depth $L$, which gave rise to the emergent scale, $\gamma=L/n^{1-s}$, that also controls the hierarchical structure of dynamical equations for wide and deep neural networks.

Theoretically, for each combination of the width $n$, depth $L$, and metaparameter $s$, we have an effective theory that describes the distribution of dynamical observables -- at the point of initialization, at the end of training, and anywhen in between -- for finite-width neural networks~\cite{PDLT}. Taking a bird's-eye view on all of them, we now see the web of effective theories, with each thread -- indexed by $(n,L)$ and parametrized by $s$ -- smoothly connecting the NT scaling at $s=0$ and the MU scaling at $s=1$.\footnote{Metaphorically, different architectures give rise to different webs. Also, to keep forcing this metaphor, each point on the web has internal dimensions, e.g., parametrized by order-one hyperparameters such as $C_{b,W}^{(\ell)}$ and $\lambda_{b,W}^{(\ell)}$.}

Formally, to know what each scaling strategy would be doing for excessively large-scale models -- and to make contact with the literature -- it is instructive to take various infinite-model-size limits.
\bi
\item The frozen NTK limit~\cite{jacot2018neural}: $0\leq s<1$; $n\rightarrow\infty$; $L=\text{fixed}$.\\
For the NT scaling strategy with $s=0$ -- and for any $s\in(0,1)$ -- taking the infinite-width limit $n\rightarrow\infty$ with fixed depth results in the vanishing $\gamma\rightarrow0$. In fact, the training dynamics of a neural network reduce to those of a linear model with an infinite number of random frozen features, resulting in kernel-machine predictions at the end of training~\cite{lee2019wide}. In particular, while its dynamics are free and solvable, there is no representation learning~\cite{chizat2019lazy,geiger2020disentangling,PDLT}.

\item The infinite-width-and-depth NT limit~\cite{hanin2020products,hanin2019finite,hu2021random,li2021future,li2022neural}: $s=0$; $n,L\rightarrow\infty$; $L/n=\text{fixed}>0$.\\
The scale $\gamma$ stays positive and inflating neural networks can keep learning representation from data.\footnote{More specifically, the contributions from the dNTK, ddNTKs, and NTK fluctuations all scale with $L/n=\gamma$. With increasing $\gamma$, there is actually a tradeoff between the benefit of representation learning and the cost of instantiation-to-instantiation fluctuations, and this tradeoff leads to the notion of the optimal aspect ratio, $\gamma^{\star}$~\cite{PDLT}.} In addition, when the depth-to-width aspect ratio $\gamma=L/n$ of a network is small, its dynamics become weakly-coupled and can be solved perturbatively~\cite{dyer2019asymptotics,PDLT}.

\item The mean-field~\cite{mei2018mean,rotskoff2018trainability,sirignano2020mean1,chizat2018global} and MU~\cite{yang2021tensor} limit: $s=1$; $n\rightarrow\infty$; $L=\text{fixed}>1$.\\
The scale $\gamma$ is of order one and the dynamics in the resulting (non)effective theory description are strongly-coupled: Taylor expansions cannot be perturbatively truncated and the infinitudes of dynamical observables get coupled.
This calls for new theoretical approaches.\footnote{See, e.g., Ref.~\cite{bordelon2022self} for an approach inspired by dynamical mean field theory.
There could also be some strong-coupling expansion or, better yet, some duality: modern field theorists don't even blink at such highly-speculative statements.}

\item A missing limit: $0<s<1$; $n,L\rightarrow\infty$; $L/n^{1-s}=\text{fixed}>0$.\\
This limit sits between the previous two limits.  In the strict limit, this family degenerates, in that the value of $s$ is irrelevant and the limit is completely parametrized by $\gamma$.\footnote{With that said, at finite width, we should stress that the metaparameter $s$ \textit{does} affect the degree of separation between two scales, $\frac{L}{n^{1-s}}$ and $\frac{L}{n}$, which might have practical consequences.} This limit is different from the infinite-width-and-depth NT limit with $s=0$ because observables proportional to $O\le(L/n\ri)$ vanish and those proportional to $O\le(L/n^{1-s}\ri)$ survive, while both survived for $s=0$. It is also different from the mean-field/MU limit with $s=1$ because we have a continuous knob $\gamma$,  while we had only a discrete knob $L$ for $s=1$. In particular, the resulting dynamics again become weakly-coupled and perturbatively analyzable for small $\gamma$.

\ei

Practically, given those limiting behaviors, it appears prudent to scale up the model with fixed $\gamma=L/n^{1-s}$ as long as tasks at hand benefit from representation learning. Beyond that, our theoretical understanding is yet incomplete to predict how each scaling strategy fairs against others -- when and why -- and for the time being the verdict is still out, both theoretically and empirically. But -- if we may muse -- just as the nature is described by weakly-coupled QED at one scale and strongly-coupled QCD at another scale, it is hard to imagine that one singular scaling strategy would be superior to all the others for all the problems.

\section*{Acknowledgements}
The author is variously grateful to Mario Geiger, Andrey Gromov, Boris Hanin,
Dan Roberts, Kushal Tirumala, and a duo of
Bruno De Luca and Eva Silverstein for discussions with varying degrees of enlightenment, and
to Yasaman Bahri, Surya Ganguli, Jaehoon Lee,
Mitchell Wortsman, and Susan Zhang for varying degrees of
positive reinforcement to write and put out this note.

\
\\
\textit{Addendum}: after the initial posting of this note, it was kindly brought to our attention by Greg Yang that the one-parameter family discussed herein is equivalent to the ``uniform parametrization'' discussed in Appendix G (Theorem G.4) of Ref.~\cite{yang2021tensor}. Our main thesis remains intact: all the scaling strategies in this family -- not just the MU scaling strategy -- enable representation learning for large-scale neural networks, as long as the depth is scaled properly with the width.

\bibliographystyle{utphys}
\bibliography{NTMU}{}

\providecommand{\href}[2]{#2}\begingroup\raggedright\begin{thebibliography}{10}

\bibitem{ramesh2021zero}
A.~Ramesh, M.~Pavlov, G.~Goh, S.~Gray, C.~Voss, A.~Radford, M.~Chen, and
  I.~Sutskever, ``{Z}ero-{S}hot {T}ext-to-{I}mage {G}eneration,'' in {\em
  International Conference on Machine Learning}, pp.~8821--8831.
\newblock 2021.

\bibitem{ramesh2022hierarchical}
A.~Ramesh, P.~Dhariwal, A.~Nichol, C.~Chu, and M.~Chen, ``Hierarchical
  {T}ext-{C}onditional {I}mage {G}eneration with {CLIP} {L}atents,''
  \href{http://arxiv.org/abs/2204.06125}{{\ttfamily arXiv:2204.06125 [cs.CV]}}.

\bibitem{costa2022no}
M.~R. Costa-juss{\`a}, J.~Cross, O.~{\c{C}}elebi, M.~Elbayad, K.~Heafield,
  K.~Heffernan, E.~Kalbassi, J.~Lam, D.~Licht, J.~Maillard, {\em et~al.},
  ``{N}o {L}anguage {L}eft {B}ehind: {S}caling {H}uman-{C}entered {M}achine
  {T}ranslation,'' \href{http://arxiv.org/abs/2207.04672}{{\ttfamily
  arXiv:2207.04672 [cs.CL]}}.

\bibitem{turing1950computing}
A.~M. Turing, ``{{C}omputing {M}achinery and {I}ntelligence},'' {\em Mind}
  {\bfseries LIX} no.~236, (1950) 433--460.

\bibitem{srivastava2022beyond}
A.~Srivastava, A.~Rastogi, A.~Rao, A.~A.~M. Shoeb, A.~Abid, A.~Fisch, A.~R.
  Brown, A.~Santoro, A.~Gupta, A.~Garriga-Alonso, {\em et~al.}, ``{B}eyond the
  {I}mitation {G}ame: {Q}uantifying and extrapolating the capabilities of
  language models,'' \href{http://arxiv.org/abs/2206.04615}{{\ttfamily
  arXiv:2206.04615 [cs.CL]}}.

\bibitem{hestness2017deep}
J.~Hestness, S.~Narang, N.~Ardalani, G.~Diamos, H.~Jun, H.~Kianinejad, M.~M.~A.
  Patwary, Y.~Yang, and Y.~Zhou, ``{D}eep {L}earning {S}caling is
  {P}redictable, {E}mpirically,''
  \href{http://arxiv.org/abs/1712.00409}{{\ttfamily arXiv:1712.00409 [cs.LG]}}.

\bibitem{kaplan2020scaling}
J.~Kaplan, S.~McCandlish, T.~Henighan, T.~B. Brown, B.~Chess, R.~Child,
  S.~Gray, A.~Radford, J.~Wu, and D.~Amodei, ``Scaling {L}aws for {N}eural
  {L}anguage {M}odels,'' \href{http://arxiv.org/abs/2001.08361}{{\ttfamily
  arXiv:2001.08361 [cs.LG]}}.

\bibitem{hoffmann2022training}
J.~Hoffmann, S.~Borgeaud, A.~Mensch, E.~Buchatskaya, T.~Cai, E.~Rutherford,
  D.~d.~L. Casas, L.~A. Hendricks, J.~Welbl, A.~Clark, {\em et~al.},
  ``{T}raining {C}ompute-{O}ptimal {L}arge {L}anguage {M}odels,''
  \href{http://arxiv.org/abs/2203.15556}{{\ttfamily arXiv:2203.15556 [cs.CL]}}.

\bibitem{jacot2018neural}
A.~Jacot, F.~Gabriel, and C.~Hongler, ``{N}eural {T}angent {K}ernel:
  {C}onvergence and {G}eneralization in {N}eural {N}etworks,'' in {\em Advances
  in Neural Information Processing Systems}, vol.~31, pp.~8571--8580.
\newblock 2018.

\bibitem{Neal1996}
R.~M. Neal, ``{P}riors for {I}nfinite {N}etworks,'' in {\em Bayesian Learning
  for Neural Networks}, pp.~29--53.
\newblock Springer, 1996.

\bibitem{LBNSPS2017}
J.~Lee, Y.~Bahri, R.~Novak, S.~S. Schoenholz, J.~Pennington, and
  J.~Sohl-Dickstein, ``{D}eep {N}eural {N}etworks as {G}aussian {P}rocesses,''
  in {\em International Conference on Learning Representations}.
\newblock 2018.

\bibitem{MRHTG2018}
A.~G.~d.~G. Matthews, M.~Rowland, J.~Hron, R.~E. Turner, and Z.~Ghahramani,
  ``{G}aussian {P}rocess {B}ehaviour in {W}ide {D}eep {N}eural {N}etworks,'' in
  {\em International Conference on Learning Representations}.
\newblock 2018.

\bibitem{yang2021tensor}
G.~Yang and E.~J. Hu, ``{T}ensor {P}rograms {IV}: {F}eature {L}earning in
  {I}nfinite-{W}idth {N}eural {N}etworks,'' in {\em International Conference on
  Machine Learning}, pp.~11727--11737.
\newblock 2021.

\bibitem{mei2018mean}
S.~Mei, A.~Montanari, and P.-M. Nguyen, ``{A} mean field view of the landscape
  of two-layer neural networks,'' {\em Proceedings of the National Academy of
  Sciences} {\bfseries 115} no.~33, (2018) E7665--E7671.

\bibitem{rotskoff2018trainability}
G.~M. Rotskoff and E.~Vanden-Eijnden, ``{T}rainability and {A}ccuracy of
  {N}eural {N}etworks: {A}n {I}nteracting {P}article {S}ystem {A}pproach,''
  \href{http://arxiv.org/abs/1805.00915}{{\ttfamily arXiv:1805.00915
  [stat.ML]}}.

\bibitem{sirignano2020mean1}
J.~Sirignano and K.~Spiliopoulos, ``{M}ean {F}ield {A}nalysis of {N}eural
  {N}etworks: {A} {L}aw of {L}arge {N}umbers,'' {\em SIAM Journal on Applied
  Mathematics} {\bfseries 80} no.~2, (2020) 725--752.

\bibitem{chizat2018global}
L.~Chizat and F.~Bach, ``{O}n the {G}lobal {C}onvergence of {G}radient
  {D}escent for {O}ver-parameterized {M}odels using {O}ptimal {T}ransport,'' in
  {\em Advances in Neural Information Processing Systems}, vol.~31,
  pp.~3036--3046.
\newblock 2018.

\bibitem{yang2022tensor}
G.~Yang, E.~J. Hu, I.~Babuschkin, S.~Sidor, X.~Liu, D.~Farhi, N.~Ryder,
  J.~Pachocki, W.~Chen, and J.~Gao, ``Tensor {P}rograms {V}: {T}uning {L}arge
  {N}eural {N}etworks via {Z}ero-{S}hot {H}yperparameter {T}ransfer,''
  \href{http://arxiv.org/abs/2203.03466}{{\ttfamily arXiv:2203.03466 [cs.LG]}}.

\bibitem{PDLT}
D.~A. Roberts, S.~Yaida, and B.~Hanin, {\em The Principles of Deep Learning
  Theory}.
\newblock Cambridge University Press, 2022.
\newblock \url{https://deeplearningtheory.com}.

\bibitem{dyer2019asymptotics}
E.~Dyer and G.~Gur-Ari, ``{A}symptotics of {W}ide {N}etworks from {F}eynman
  {D}iagrams,'' in {\em International Conference on Learning Representations}.
\newblock 2020.

\bibitem{liu2022convnet}
Z.~Liu, H.~Mao, C.-Y. Wu, C.~Feichtenhofer, T.~Darrell, and S.~Xie, ``{A}
  {C}onv{N}et for the 2020s,'' in {\em IEEE/CVF Conference on Computer Vision
  and Pattern Recognition}, pp.~11976--11986.
\newblock 2022.

\bibitem{poole2016exponential}
B.~Poole, S.~Lahiri, M.~Raghu, J.~Sohl-Dickstein, and S.~Ganguli,
  ``{E}xponential expressivity in deep neural networks through transient
  chaos,'' in {\em Advances in Neural Information Processing Systems}, vol.~29,
  pp.~3360--3368.
\newblock 2016.

\bibitem{raghu2017expressive}
M.~Raghu, B.~Poole, J.~Kleinberg, S.~Ganguli, and J.~Sohl-Dickstein, ``{O}n the
  {E}xpressive {P}ower of {D}eep {N}eural {N}etworks,'' in {\em International
  Conference on Machine Learning}, pp.~2847--2854.
\newblock 2017.

\bibitem{schoenholz2016deep}
S.~S. Schoenholz, J.~Gilmer, S.~Ganguli, and J.~Sohl-Dickstein, ``{D}eep
  {I}nformation {P}ropagation,'' in {\em International Conference on Learning
  Representations}.
\newblock 2017.

\bibitem{glorot2010understanding}
X.~Glorot and Y.~Bengio, ``{U}nderstanding the difficulty of training deep
  feedforward neural networks,'' in {\em International Conference on Artificial
  Intelligence and Statistics}, pp.~249--256.
\newblock 2010.

\bibitem{he2015delving}
K.~He, X.~Zhang, S.~Ren, and J.~Sun, ``{D}elving {D}eep into {R}ectifiers:
  {S}urpassing {H}uman-{L}evel {P}erformance on {I}mage{N}et
  {C}lassification,'' in {\em IEEE International Conference on Computer
  Vision}, pp.~1026--1034.
\newblock 2015.

\bibitem{coleman1967all}
S.~Coleman and J.~Mandula, ``All {P}ossible {S}ymmetries of the ${S}$
  {M}atrix,'' {\em Physical Review} {\bfseries 159} no.~5, (1967) 1251.

\bibitem{weinberg2005quantum}
S.~Weinberg, {\em The Quantum Theory of Fields, Volume 3: Supersymmetry}.
\newblock Cambridge University Press, 2005.

\bibitem{lee2019wide}
J.~Lee, L.~Xiao, S.~Schoenholz, Y.~Bahri, R.~Novak, J.~Sohl-Dickstein, and
  J.~Pennington, ``{W}ide {N}eural {N}etworks of {A}ny {D}epth {E}volve as
  {L}inear {M}odels {U}nder {G}radient {D}escent,'' in {\em Advances in Neural
  Information Processing Systems}, vol.~32, pp.~8572--8583.
\newblock 2019.

\bibitem{chizat2019lazy}
L.~Chizat, E.~Oyallon, and F.~Bach, ``{O}n {L}azy {T}raining in
  {D}ifferentiable {P}rogramming,'' in {\em Advances in Neural Information
  Processing Systems}, vol.~32, pp.~2937--2947.
\newblock 2019.

\bibitem{geiger2020disentangling}
M.~Geiger, S.~Spigler, A.~Jacot, and M.~Wyart, ``{D}isentangling feature and
  lazy training in deep neural networks,'' {\em Journal of Statistical
  Mechanics: Theory and Experiment} {\bfseries 2020} no.~11, (2020) 113301.

\bibitem{hanin2020products}
B.~Hanin and M.~Nica, ``Products of {M}any {L}arge {R}andom {M}atrices and
  {G}radients in {D}eep {N}eural {N}etworks,'' {\em Communications in
  Mathematical Physics} {\bfseries 376} no.~1, (2020) 287--322.

\bibitem{hanin2019finite}
B.~Hanin and M.~Nica, ``{F}inite {D}epth and {W}idth {C}orrections to the
  {N}eural {T}angent {K}ernel,'' in {\em International Conference on Learning
  Representations}.
\newblock 2019.

\bibitem{hu2021random}
Z.~Hu and H.~Huang, ``{O}n the {R}andom {C}onjugate {K}ernel and {N}eural
  {T}angent {K}ernel,'' in {\em International Conference on Machine Learning},
  pp.~4359--4368.
\newblock 2021.

\bibitem{li2021future}
M.~Li, M.~Nica, and D.~Roy, ``{T}he future is log-{G}aussian: {R}es{N}ets and
  their infinite-depth-and-width limit at initialization,'' in {\em Advances in
  Neural Information Processing Systems}, vol.~34, pp.~7852--7864.
\newblock 2021.

\bibitem{li2022neural}
M.~B. Li, M.~Nica, and D.~M. Roy, ``{T}he {N}eural {C}ovariance {SDE}: {S}haped
  {I}nfinite {D}epth-and-{W}idth {N}etworks at {I}nitialization,''
  \href{http://arxiv.org/abs/2206.02768}{{\ttfamily arXiv:2206.02768
  [stat.ML]}}.

\bibitem{bordelon2022self}
B.~Bordelon and C.~Pehlevan, ``{S}elf-{C}onsistent {D}ynamical {F}ield {T}heory
  of {K}ernel {E}volution in {W}ide {N}eural {N}etworks,''
  \href{http://arxiv.org/abs/2205.09653}{{\ttfamily arXiv:2205.09653
  [stat.ML]}}.

\bibitem{aitken2020asymptotics}
K.~Aitken and G.~Gur-Ari, ``{O}n the asymptotics of wide networks with
  polynomial activations,'' \href{http://arxiv.org/abs/2006.06687}{{\ttfamily
  arXiv:2006.06687 [cs.LG]}}.

\bibitem{huang2020dynamics}
J.~Huang and H.-T. Yau, ``{D}ynamics of {D}eep {N}eural {N}etworks and {N}eural
  {T}angent {H}ierarchy,'' in {\em International Conference on Machine
  Learning}, pp.~4542--4551.
\newblock 2020.

\end{thebibliography}\endgroup

\newpage

\appendix

\section{Peeking into the Hierarchical Structure}\label{app:hierarchy}
In this Appendix, we'll compute the higher-order differentials of the NTK for one-hidden-layer neural networks. We'll do this in two steps: in \S\ref{subsec:onehidden-typeI} we'll explicitly calculate \textit{type-I} differentials to see the hierarchical pattern; then, knowing the pattern, in \S\ref{subsec:onehidden-general} we'll implicitly show by induction that the same hierarchical structure holds for general types of the higher-order differentials.

Beyond seeing the hierarchical structure with the general metaparameter $s$, this Appendix serves two additional purposes: for one, it complements the existing work that estimates the width scalings of these higher-order differentials for deeper neural networks with a \texttt{linear}~\cite{dyer2019asymptotics} or \texttt{polynomial}~\cite{aitken2020asymptotics} activation function (see also Ref.~\cite{huang2020dynamics} for various bounds on some combinations of higher-order differentials); for two, it might help someone in the future trying to show the same hierarchical structure for deeper neural networks with general activation functions, using the result and derivation herein as stepping stones for the recursive analysis.

Before diving into the actual calculations, let us make three preparatory general remarks.
\bi
\item Thinking about the way in which higher-order differentials of the NTK are being generated, we can recognize that the $k$-th-order differentials can be generated by acting on a set of the $(k-1)$-th-order differentials with the differential operator
\be\label{eq:generating-differential-operator}
\sum_{\mu\nu}\le(\frac{d\out_{i;\delta}}{d\theta_{\nu}}\ri)\lambda_{\mu\nu}\frac{d}{d\theta_{\mu}}\, .
\ee
For instance, acting on the NTK with the operator generates the dNTK as
\begin{align}
&\sum_{\mu_2\nu_2}\le(\frac{d\out_{i_2;\delta_2}}{d\theta_{\nu_2}}\ri)\lambda_{\mu_2\nu_2}\frac{d}{d\theta_{\mu_2}}\le[\sum_{\mu_1,\nu_1}\lambda_{\mu_1\nu_1}\frac{d\out_{i_0;\delta_0}}{d\theta_{\mu_1}}\frac{d\out_{i_1;\delta_1}}{d\theta_{\nu_1}}\ri]\, \\
=&\sum \lambda_{\mu_1\nu_1}\lambda_{\mu_2\nu_2}\le(\frac{d^2\out_{i_0;\delta_0}}{d\theta_{\mu_1}d\theta_{\mu_2}}\frac{d\out_{i_1;\delta_1}}{d\theta_{\nu_1}}\frac{d\out_{i_2;\delta_2}}{d\theta_{\nu_2}}+\frac{d\out_{i_0;\delta_0}}{d\theta_{\mu_1}}\frac{d^2\out_{i_1;\delta_1}}{d\theta_{\nu_1}d\theta_{\mu_2}}\frac{d\out_{i_2;\delta_2}}{d\theta_{\nu_2}}\ri)\, \notag\\
=&\mathrm{d}H_{i_0i_1i_2;\delta_0\delta_1\delta_2}+\mathrm{d}H_{i_1i_0i_2;\delta_1\delta_0\delta_2}\, ,\notag
\end{align}
and acting on the dNTK with the operator generates the two types of the ddNTKs as
\begin{align}
&\sum_{\mu_3\nu_3}\le(\frac{d\out_{i_3;\delta_3}}{d\theta_{\nu_3}}\ri)\lambda_{\mu_3\nu_3}\frac{d}{d\theta_{\mu_3}}\le[\sum_{\mu_1,\nu_1,\mu_2,\nu_2}\lambda_{\mu_1\nu_1}\lambda_{\mu_2\nu_2}\frac{d^2\out_{i_0;\delta_0}}{d\theta_{\mu_1}d\theta_{\mu_2}}\frac{d\out_{i_1;\delta_1}}{d\theta_{\nu_1}}\frac{d\out_{i_2;\delta_2}}{d\theta_{\nu_2}}\ri]\, \\
=&\sum \lambda_{\mu_1\nu_1}\lambda_{\mu_2\nu_2}\lambda_{\mu_3\nu_3}\bigg(\frac{d^3\out_{i_0;\delta_0}}{d\theta_{\mu_1}d\theta_{\mu_2}d\theta_{\mu_3}}\frac{d\out_{i_1;\delta_1}}{d\theta_{\nu_1}}\frac{d\out_{i_2;\delta_2}}{d\theta_{\nu_2}}\frac{d\out_{i_3;\delta_3}}{d\theta_{\nu_3}}\, \notag\\
&\ \ \ \ \ \ \ \ \ \ \ \ \ \ \ \ \ \ \ \ \ \ \ \ \ \ \ \ +\frac{d^2\out_{i_0;\delta_0}}{d\theta_{\mu_1}d\theta_{\mu_2}}\frac{d^2\out_{i_1;\delta_1}}{d\theta_{\nu_1}d\theta_{\mu_3}}\frac{d\out_{i_2;\delta_2}}{d\theta_{\nu_2}}\frac{d\out_{i_3;\delta_3}}{d\theta_{\nu_3}}+\frac{d^2\out_{i_0;\delta_0}}{d\theta_{\mu_1}d\theta_{\mu_2}}\frac{d\out_{i_1;\delta_1}}{d\theta_{\nu_1}}\frac{d^2\out_{i_2;\delta_2}}{d\theta_{\nu_2}d\theta_{\mu_3}}\frac{d\out_{i_3;\delta_3}}{d\theta_{\nu_3}}\bigg)\, \notag\\
=&\mathrm{dd}_{\mathrm{I}}H_{i_0i_1i_2i_3;\delta_0\delta_1\delta_2\delta_3}+\mathrm{dd}_{\mathrm{II}}H_{i_0i_1i_2i_3;\delta_0\delta_1\delta_2\delta_3}+\mathrm{dd}_{\mathrm{II}}H_{i_0i_2i_1i_3;\delta_0\delta_2\delta_1\delta_3}\, .\notag
\end{align}
\item Among various types of the higher-order differentials, we single out those with the form
\be
\mathrm{d}^{k-1}_{\mathrm{I}}H_{i_0\cdots i_{k};\delta_0\cdots\delta_{k}}\equiv \sum_{\substack{\mu_1,\nu_1\\ \cdots\\ \mu_{k},\nu_{k}}}\lambda_{\mu_1\nu_1}\cdots\lambda_{\mu_{k}\nu_{k}}\frac{d^{k}\out_{i_0;\delta_0}}{d\theta_{\mu_1}\cdots d\theta_{\mu_{k}}}\frac{d\out_{i_1;\delta_1}}{d\theta_{\nu_1}}\cdots\frac{d\out_{i_{k};\delta_{k}}}{d\theta_{\nu_{k}}}\, \label{eq:ddddINTKdef}
\ee
as the \textit{type-I} $(k-1)$-th-order differentials of the NTK.  These are the ones that appear in the dynamical update equation for the network outputs~\eqref{eq:output-update}.
\item We decompose the learning-rate tensor into those that act within each layer as 
\be
\lambda_{\mu\nu}=\sum_{\ell=1}^{L}\lambda_{\mu\nu}^{(\ell)}\, ,\label{eq:lambda-decomposition}
\ee
and, with a slight abuse of notation, denote the model parameters in the $\ell$-th layer as $\theta^{(\ell)}_{\mu}$.
\ei

Now, let's dive in.

\subsection{Type-I Higher-Order Differentials}\label{subsec:onehidden-typeI}
First, let's write out the type-I $(k-1)$-th-order differential~\eqref{eq:ddddINTKdef} for the one-hidden-layer neural networks (i.e., those with $L=2$), with the decomposition~\eqref{eq:lambda-decomposition} in mind:
\be
\widehat{\mathrm{d}^{k-1}_{\mathrm{I}}H}_{i_0\cdots i_{k};\delta_0\cdots\delta_{k}}=\sum_{\ell_1,\ldots,\ell_k=1}^{2}\sum_{\substack{\mu_1,\nu_1\\ \cdots\\ \mu_{k},\nu_{k}}}\lambda_{\mu_1\nu_1}^{(\ell_1)}\cdots\lambda_{\mu_{k}\nu_{k}}^{(\ell_k)}\frac{d^{k}z^{(2)}_{i_0;\delta_0}}{d\theta_{\mu_1}^{(\ell_1)}\cdots d\theta_{\mu_{k}}^{(\ell_k)}}\frac{dz^{(2)}_{i_1;\delta_1}}{d\theta_{\nu_1}^{(\ell_1)}}\cdots\frac{dz^{(2)}_{i_{k};\delta_{k}}}{d\theta_{\nu_{k}}^{(\ell_k)}}\, .
\ee
Here, for one last time, we've put a hat to remind us that we are computing this object at initialization, which can be used to read off the scalings and statistics after training as well if wished. There are two kinds of nonzero terms in the sum over layers, the first kind with $\ell_1=\cdots=\ell_k=1$ and the second kind with $(\ell_1,\ell_2,\ldots,\ell_k)=(2,1,\ldots,1)$ along with its permutations thereof. All the other terms vanish as the output $z^{(2)}_{i_0;\delta_0}$ is linear in the second-layer model parameters $\theta_{\mu}^{(2)}$.

To evaluate the first kind, note that
\be
\frac{dz^{(2)}_{i;\delta}}{d\theta_{\nu}^{(1)}}=\sum_{j=1}^{n_1}\frac{dz^{(2)}_{i;\delta}}{dz_{j;\delta}^{(1)}}\frac{dz^{(1)}_{j;\delta}}{d\theta_{\nu}^{(1)}}=\sum_{j=1}^{n_1}W^{(2)}_{ij}\sigma'\le(z_{j;\delta}^{(1)}\ri)\frac{dz^{(1)}_{j;\delta}}{d\theta_{\nu}^{(1)}}\, 
\ee
by the chain rule, and similarly -- using the fact that the first-layer preactivations $z^{(1)}_{i;\delta}$ are linear in the first-layer model parameters $\theta_{\mu}^{(1)}$ -- note that
\be
\frac{d^{k}z^{(2)}_{i_0;\delta_0}}{d\theta_{\mu_1}^{(1)}\cdots d\theta_{\mu_{k}}^{(1)}}=\sum_{j_0=1}^{n_1}W^{(2)}_{i_0j_0}\sigma^{[k]}\le(z_{j_0;\delta_0}^{(1)}\ri)\frac{dz^{(1)}_{j_0;\delta_0}}{d\theta_{\mu_1}^{(1)}}\cdots \frac{dz^{(1)}_{j_0;\delta_0}}{d\theta_{\mu_k}^{(1)}}\, ,
\ee
where we denoted the $k$-th derivative of the activation function by $\sigma^{[k]}$. Putting them together, we get the contribution
\begin{align}
&\sum_{\substack{\mu_1,\nu_1\\ \cdots\\ \mu_{k},\nu_{k}}}\lambda_{\mu_1\nu_1}^{(1)}\cdots\lambda_{\mu_{k}\nu_{k}}^{(1)}
\sum_{j_0,j_1,\ldots,j_k=1}^{n_1}W^{(2)}_{i_0j_0}W^{(2)}_{i_1j_1}\cdots W^{(2)}_{i_kj_k}
\sigma^{[k]}\le(z_{j_0;\delta_0}^{(1)}\ri)\sigma'\le(z_{j_1;\delta_1}^{(1)}\ri)\cdots\sigma'\le(z_{j_k;\delta_k}^{(1)}\ri)\, \notag\\
&\ \ \ \ \ \ \ \ \ \ \ \ \ \ \ \ \ \ \ \ \ \ \ \ \ \ \ \ \ \ \ \ \ \ \ \ \ \ \times \frac{dz^{(1)}_{j_0;\delta_0}}{d\theta_{\mu_1}^{(1)}}\cdots \frac{dz^{(1)}_{j_0;\delta_0}}{d\theta_{\mu_k}^{(1)}}\frac{dz^{(1)}_{j_1;\delta_1}}{d\theta_{\nu_1}^{(1)}}\cdots\frac{dz^{(1)}_{j_k;\delta_k}}{d\theta_{\nu_k}^{(1)}}\,\notag \\
=&\sum_{j_0,j_1,\ldots,j_k=1}^{n_1}W^{(2)}_{i_0j_0}W^{(2)}_{i_1j_1}\cdots W^{(2)}_{i_kj_k}
\sigma^{[k]}\le(z_{j_0;\delta_0}^{(1)}\ri)\sigma'\le(z_{j_1;\delta_1}^{(1)}\ri)\cdots\sigma'\le(z_{j_k;\delta_k}^{(1)}\ri) 
\NTK^{(1)}_{j_0j_1;\delta_0\delta_1}\cdots  \NTK^{(1)}_{j_0j_k;\delta_0\delta_k}\, \notag\\
=&\sum_{j=1}^{n_1}W^{(2)}_{i_0j}W^{(2)}_{i_1j}\cdots W^{(2)}_{i_kj}
\sigma^{[k]}\le(z_{j;\delta_0}^{(1)}\ri)\sigma'\le(z_{j;\delta_1}^{(1)}\ri)\cdots\sigma'\le(z_{j;\delta_k}^{(1)}\ri) H^{(1)}_{\delta_0\delta_1}\cdots  H^{(1)}_{\delta_0\delta_k}\, ,\label{eq:ddddNTKfirstkind}
\end{align}
where in the last line we used the fact that the first-layer NTK~\eqref{eq:firstNTK},
\be
\NTK^{(1)}_{j_0j_1;\delta_0\delta_1}=\frac{1}{n^{q_1}}\le[\Lb^{(1)}+\LW^{(1)}\le(\frac{1}{n_0}\sum_{j=1}^{n_0}x_{j;\delta_0}x_{j;\delta_1}\ri)\ri]\delta_{j_0j_1}\equiv\delta_{j_0j_1} H^{(1)}_{\delta_0\delta_1}, \, 
\ee
is diagonal in neural indices $i$.

To evaluate the second kind, note that
\be
\frac{dz^{(2)}_{i_1;\delta_1}}{dW^{(2)}_{ij}}=\delta_{i_1i}\sigma\le(z^{(1)}_{j;\delta_1}\ri)\, ,
\ee
and
\be
\frac{d^{k}z^{(2)}_{i_0;\delta_0}}{dW^{(2)}_{ij}d\theta_{\mu_2}^{(1)}\cdots d\theta_{\mu_{k}}^{(1)}}=\delta_{i_0i}\sigma^{[k-1]}\le(z_{j;\delta_0}^{(1)}\ri)\frac{dz^{(1)}_{j;\delta_0}}{d\theta_{\mu_2}^{(1)}}\cdots \frac{dz^{(1)}_{j;\delta_0}}{d\theta_{\mu_k}^{(1)}}\, .
\ee
Putting them together  -- summing over all the components of the second-layer weights and noting that there is no second-layer bias contribution for $(k-1)\geq1$ --  for  the $(\ell_1,\ell_2,\ldots,\ell_k)=(2,1,\ldots,1)$, we get the contribution
\begin{align}
&\frac{1}{n^{q_2}}\frac{\lambda^{(2)}_W}{n_1}\delta_{i_0i_1}\sum_{\substack{\mu_2,\nu_2\\ \cdots\\ \mu_{k},\nu_{k}}}\lambda_{\mu_2\nu_2}^{(1)}\cdots\lambda_{\mu_{k}\nu_{k}}^{(1)}\sum_{j,j_2,\ldots,j_k=1}^{n_1}W^{(2)}_{i_2j_2}\cdots W^{(2)}_{i_kj_k}\, \notag\\
&\ \ \times
\sigma^{[k-1]}\le(z_{j;\delta_0}^{(1)}\ri)\sigma\le(z_{j;\delta_1}^{(1)}\ri)\sigma'\le(z_{j_2;\delta_2}^{(1)}\ri)\cdots\sigma'\le(z_{j_k;\delta_k}^{(1)}\ri)\frac{dz^{(1)}_{j;\delta_0}}{d\theta_{\mu_2}^{(1)}}\cdots \frac{dz^{(1)}_{j;\delta_0}}{d\theta_{\mu_k}^{(1)}}\frac{dz^{(1)}_{j_2;\delta_2}}{d\theta_{\nu_2}^{(1)}}\cdots\frac{dz^{(1)}_{j_k;\delta_k}}{d\theta_{\nu_k}^{(1)}}\, \notag\\
=&\frac{1}{n^{q_2}}\frac{\lambda^{(2)}_W}{n_1}\delta_{i_0i_1}\!\!\!\!\!\!\!\sum_{j,j_2,\ldots,j_k=1}^{n_1}\!\!\!\!\!\!\!W^{(2)}_{i_2j_2}\cdots W^{(2)}_{i_kj_k}
\sigma^{[k-1]}\!\le(z_{j;\delta_0}^{(1)}\ri)\sigma\!\le(z_{j;\delta_1}^{(1)}\ri)\sigma'\!\le(z_{j_2;\delta_2}^{(1)}\ri)\cdots\sigma'\!\le(z_{j_k;\delta_k}^{(1)}\ri) 
\NTK^{(1)}_{jj_2;\delta_0\delta_2}\cdots  \NTK^{(1)}_{jj_k;\delta_0\delta_k}\, \notag\\
=&\frac{1}{n^{q_2}}\delta_{i_0i_1}\frac{\lambda^{(2)}_W}{n_1}\sum_{j=1}^{n_1}W^{(2)}_{i_2j}\cdots W^{(2)}_{i_kj}
\sigma^{[k-1]}\!\le(z_{j;\delta_0}^{(1)}\ri)\sigma\!\le(z_{j;\delta_1}^{(1)}\ri)\sigma'\!\le(z_{j;\delta_2}^{(1)}\ri)\cdots\sigma'\!\le(z_{j;\delta_k}^{(1)}\ri) H^{(1)}_{\delta_0\delta_2}\cdots  H^{(1)}_{\delta_0\delta_k}\, ,\label{eq:ddddNTKsecondkind}
\end{align}
and we get similar contributions from the appropriate permutations.

Let's compute the scalings of these contributions with our meta-principled family of hyperparameter scaling strategies~\eqref{eq:ultimate-exponents}, $p_1=q_1=0$ and $p_2=q_2=r=s$. With $p_1=q_1=0$, we note that the first-layer preactivations $z^{(1)}$ and the first-layer NTK $H^{(1)}$ are both of order one, so we can essentially forget about them for a moment. With that noted, there are two cases to consider.
\bi
\item  For odd $k=2m+1$, that is, for even $k-1=2m$, let's evaluate the following expectation value of the gauge-invariant product at initialization
\be
\E{\eta^{2m+1}\widehat{\mathrm{d}^{2m}_{\mathrm{I}}H}_{i_0\cdots i_{2m+1};\delta_0\cdots\delta_{2m+1}}}\, .
\ee
For the first kind of contributions~\eqref{eq:ddddNTKfirstkind}, we get a factor of $1/n^{(m+1) (p_2+1)}=1/n^{(m+1)(s+1)}$ from $(2m+2)$ weights while it gets a factor of $n$ from the summation over $j$, so when multiplied by $\eta^{2m+1}\sim n^{2ms+s}$, we get  $O\le(1/n^{m(1-s)}\ri)$. For the second kind of contributions~\eqref{eq:ddddNTKsecondkind}, we get a factor of $1/n^{q_2}=1/n^s$, a cancelation between the denominator and summation $(1/n_1)\sum_{j=1}^{n_1}$, a factor of $1/n^{m(p_2+1)}=1/n^{m(s+1)}$ from $2m$ weights, and a factor of $\eta^{2m+1}\sim n^{2ms+s}$, so overall we once again get $O\le(1/n^{m(1-s)}\ri)$. In summary,
\be
\E{\eta^{2m+1}\widehat{\mathrm{d}^{2m}_{\mathrm{I}}H}_{i_0\cdots i_{2m+1};\delta_0\cdots\delta_{2m+1}}}=O\le(\frac{1}{n^{m(1-s)}}\ri)=O\le(\gamma^m\ri)\, 
\ee
with $\gamma\equiv 1/n^{1-s}$ -- or $\gamma=2/n^{1-s}$ if two means anything here (it doesn't). Thus, we see the hierarchical structure of higher-order differentials advertised in the main text.
\item For even $k=2m$, that is, for odd $k-1=2m-1$, let's evaluate the following expectation value of the gauge-invariant product at initialization
\be
\E{\eta^{2m}z_{i;\delta}^{(2)}\widehat{\mathrm{d}^{2m-1}_{\mathrm{I}}H}_{i_0\cdots i_{2m};\delta_0\cdots\delta_{2m}}}\, .
\ee
Note that we inserted one factor of the second-layer preactivations because -- just as we saw for the dNTK -- the expectation would otherwise vanish due to a dangling factor of the second-layer weights. With that factor added, for the first kind of contributions~\eqref{eq:ddddNTKfirstkind}, we get a factor of $1/n^{(m+1) (p_2+1)}=1/n^{(m+1)(s+1)}$ from $(2m+1+1)$ weights while it gets a factor of $n$ from the summation over $j$, so when multiplied by $\eta^{2m}\sim n^{2ms}$, we get  $O\le(1/n^{m(1-s)+s}\ri)$. Running the similar analysis, we get the same order for the second kind of contributions~\eqref{eq:ddddNTKsecondkind}. In summary
\be
\E{\eta^{2m}z_{i;\delta}^{(2)}\widehat{\mathrm{d}^{2m-1}_{\mathrm{I}}H}_{i_0\cdots i_{2m};\delta_0\cdots\delta_{2m}}}=O\le(\frac{1}{n^{m(1-s)+s}}\ri)=O\le(\gamma^m\frac{1}{n^s}\ri)=O\le(\gamma^{m-1}\frac{1}{n}\ri)\, .
\ee
We see that, for $s>0$, the contributions from the odd-order differentials are more suppressed than the ones from the even-order differentials.
\ei

\subsection{General Types of Higher-Order Differentials}\label{subsec:onehidden-general}
We'll now inductively show that the $(k-1)$-th-order differentials have the form of
\be\label{eq:inductive-form}
\le(\frac{1}{n^{q_2}}\frac{\lambda_W^{(2)}}{n_1}\ri)^{u} \sum_{j=1}^{n_1} W^{(2)}_{i_1 j} \cdots W_{i_v j}^{(2)} F\le(z^{(1)}_j, H^{(1)}\ri)\, ,
\ee
with $2u+v=k+1$ where the last factor $F\le(z^{(1)}_j, H^{(1)}\ri)$ is an order-one function of the first-layer mean NTK $H^{(1)}_{\delta_1\delta_2}$ and the \textit{$j$-th component} of the first-layer preactivation $z_{j;\delta}^{(1)}$. Note that  the two contributions~\eqref{eq:ddddNTKfirstkind}~\eqref{eq:ddddNTKsecondkind} that appeared above for the type-I differentials both respect this form.

The key to induction is to decompose the generative differential operator~\eqref{eq:generating-differential-operator} as
\begin{align}\label{eq:generating-differential-operator2}
&\sum_{\mu\nu}\le(\frac{d\out_{i;\delta}}{d\theta_{\nu}}\ri)\lambda_{\mu\nu}\frac{d}{d\theta_{\mu}}\, \\
=&\sum_{\mu\nu}\le(\frac{dz^{(2)}_{i;\delta}}{d\theta^{(1)}_{\nu}}\ri)\lambda_{\mu\nu}^{(1)}\frac{d}{d\theta_{\mu}^{(1)}}+\le(\frac{1}{n^{q_2}}\frac{\lambda_W^{(2)}}{n_1}\ri)\sum_{i'=1}^{n_2}\sum_{j'=1}^{n_1} \le(\frac{dz^{(2)}_{i;\delta}}{dW^{(2)}_{i'j'}}\ri)\frac{d}{dW^{(2)}_{i'j'}}+\le(\frac{1}{n^{q_2}}\lambda_b^{(2)}\ri)\sum_{i'=1}^{n_2}\le(\frac{dz^{(2)}_{i;\delta}}{db^{(2)}_{i'}}\ri)\frac{d}{db^{(2)}_{i'}}\, \notag\\
=&\sum_{j'=1}^{n_1}W^{(2)}_{ij'}\sigma'\le(z^{(1)}_{j';\delta}\ri)\le[\sum_{\mu\nu}\lambda_{\mu\nu}^{(1)}\frac{dz^{(1)}_{j';\delta}}{d\theta^{(1)}_{\nu}}\frac{d}{d\theta_{\mu}^{(1)}}\ri]+\le(\frac{1}{n^{q_2}}\frac{\lambda_W^{(2)}}{n_1}\ri)\sum_{j'=1}^{n_1}\sigma\le(z^{(1)}_{j';\delta}\ri)\frac{d}{dW^{(2)}_{ij'}}+\le(\frac{1}{n^{q_2}}\lambda_b^{(2)}\ri)\frac{d}{db^{(2)}_{i}}\, .\notag
\end{align}
Going from the last to the first: when the last operator hits the expression~\eqref{eq:inductive-form}, it returns zero because there is no second-layer bias; when the middle operator hits the expression~\eqref{eq:inductive-form}, it increases $u$ by one while it decreases $v$ by one -- note that the sum over $j'$ collapses onto $j$ -- so in total increases $2u+v$ by one; and when the first operator hits the expression~\eqref{eq:inductive-form}, it increases $v$ -- and hence $2u+v$ -- by one while again the sum over $j'$ collapses onto $j$ because the first-layer NTK -- which gets generated when the operator in the square parentheses hits the first-layer preactivation in the function $F$ -- is diagonal in the neural indices. So it keeps the form~\eqref{eq:inductive-form} with $2u+v=k+1$ increased by one, completing the induction.

From the expression~\eqref{eq:inductive-form}, we can run the same scaling analysis as before and find the $\gamma^m$ scaling for $2m$-th-order differentials and the $\gamma^m/n^s$ scaling for $(2m-1)$-th-order differentials.

\end{document}